\documentclass{article}

\usepackage[preprint]{neurips_2026}

\usepackage[utf8]{inputenc} 
\usepackage[T1]{fontenc}    
\usepackage{hyperref}       
\usepackage{url}            
\usepackage{booktabs}       
\usepackage{amsfonts}       
\usepackage{nicefrac}       
\usepackage{microtype}      
\usepackage{xcolor}         

\usepackage{graphicx}
\usepackage{amsmath}
\usepackage{multirow}
\usepackage[ruled,vlined]{algorithm2e}
\usepackage{listings}
\usepackage{enumitem}

\usepackage{longtable}

\usepackage[most]{tcolorbox}
\tcbuselibrary{breakable,skins}
\usepackage[dvipsnames]{xcolor}

\title{MAVEN: Multi-Agent Verification-Elaboration Network with In-Step Epistemic Auditing}

\author{%
  Yinsheng Yao\\
  Tongji University\\
  Shanghai, China \\
  \texttt{2251929@tongji.edu.cn} \\
  \And
  Jiehao Tang \\
  Tongji University \\
  Shanghai, China \\
  \texttt{2410953@tongji.edu.cn} \\
  \AND
  Zhaozhen Yang \\
  Tongji University \\
  Shanghai, China \\
  \texttt{2252712@tongji.edu.cn} \\
  \And
  Dawei Cheng $\dag$ \thanks{$\dag$ Corresponding author.} \\
  Tongji University \\
  Shanghai, China \\
  \texttt{dcheng@tongji.edu.cn} \\
}

\begin{document}

\maketitle

\begin{abstract}
While explicit reasoning trajectories enhance model interpretability, existing paradigms often rely on monolithic chains that lack intermediate verification, allowing early errors to cascade unchecked. This lack of modularity impedes granular auditing and compromises the epistemic trust required for high-stakes applications. We propose MAVEN (Multi-Agent Verification-Elaboration Network with In-Step Epistemic Auditing), a blackboard-inspired framework designed to transform LLMs into deliberate reasoners through explicit role-decoupling. At its core, MAVEN operationalizes an adversarial Skeptic-Researcher-Judge loop, simulating expert deliberation by functionally separating logical defense from factual grounding. Experiments on OpenBookQA, TruthfulQA, HALUEVAL and StrategyQA benchmarks demonstrate that MAVEN delivers superior reasoning quality across four fine-grained metrics. Notably, MAVEN consistently outperforms latent reasoning models such as GEMINI-3.1-Pro and consensus-based baselines (e.g., ReConcile) by generating explicitly structured, modular, and verifiable deliberation trajectories, rather than relying on implicit internal states or post-hoc consensus. Moreover, comprehensive evaluations confirm that MAVEN is fully model-agnostic, serving as a strong and transferable reasoning booster that yields substantial performance improvements across diverse backbone models.
\end{abstract}

\section{Introduction}
\label{sec:introduction}

Large Language Models (LLMs) have fundamentally restructured industrial workflows, particularly within knowledge-intensive sectors such as legal auditing, medical diagnostics, and automated fact-checking ~\citep{openai2023gpt4, deepseek2024v3}. In these critical contexts, the efficacy of an AI system is contingent upon the accuracy of its final output and its capacity for fact-consistent reasoning ~\citep{ji2023survey, huang2025survey}. However, as query complexity increases, the inherent constraints of monolithic reasoning architectures and linear pipelines ~\citep{wei2022chain, zhou2022least} have become pronounced. To establish epistemic trust among professionals, it is imperative to propose a multi-agent collaborative deliberation framework ~\citep{du2023improving, li2023camel} capable of providing robust, evidence-backed justifications. In the pursuit of complex multi-step reasoning, the landscape has evolved from basic Chain-of-Thought (CoT) prompting~\citep{wei2022chain, zhou2022least, wang2022self, wang2023plan}. It exposes intermediate rationales but remains brittle to error propagation. To address this problem, subsequent architectures introduced non-linear topologies such as Tree-of-Thoughts~\citep{yao2023tree} and Graph-of-Thoughts~\citep{besta2024graph, yue2023llmcascades} for richer exploration. Moreover, the paradigm has shifted toward multi-agent collaboration and consensus-based frameworks~\citep{du2023improving, li2023camel, chen2024reconcile, wang2024mixture, qian2024scaling, feng2024don}, where robustness is enhanced through iterative debate or collective orchestration~\cite{dang2025multi, liu2025breaking}. However, a persistent limitation across these structural advancements is the limited operational decoupling: while personas differ, the generation and verification stages remain tightly coupled within a monolithic process or aggregate outputs at the terminal level~\cite{wynn2025talk, li2025advancing}. Consequently, logically flawed reasoning chains can still achieve consensus through majority agreement, preserving linguistic plausibility while masking underlying factual or logical errors.

Recent frontiers in reasoning optimization have yielded substantial gains in task performance, exemplified by reinforcement-driven frameworks like DeepSeek-R1~\cite{deepseek2025r1}. Parallelly, sophisticated hallucination detection and uncertainty awareness mechanisms~\cite{ji2023survey, huang2025survey, zhao2024knowing, zhang2024rtuning, orgad2024llms, sawczyn2026factselfcheck} have been proposed to mitigate unfaithful outputs. Yet, these methods predominantly optimize for terminal accuracy or treat reasoning as an opaque latent process, offering limited guarantees on causal depth and contextual grounding. As evidenced by recent studies~\cite{kadavath2022knowledge, li2025reasoning}, correctness does not equate to trustworthiness; a correct answer derived from unsound logic is as hazardous as an explicit hallucination in mission-critical domains. Current evaluation and modeling paradigms still exhibit a critical gap in enforcing justification transparency and comparative quality throughout the reasoning trajectory~\cite{kulkarni2025evaluating, bar2025beyond}. This underscores the necessity for frameworks that transition beyond black-box inference toward an auditable, multi-stage deliberation process where every intermediate step is subject to discrete verification gates~\cite{zhang2025incentivizing}.

To address these challenges, we propose MAVEN (Multi-Agent Verification-Elaboration Network with In-Step Epistemic Auditing), a blackboard-based deliberative framework that transforms LLMs from intuitive generators into systematic reasoners. The design takes inspiration from the classic blackboard architecture ~\cite{nii1986blackboard} and supports coordinated multi-agent collaboration on a shared reasoning workspace. MAVEN is designed as a transparent and modular adversarial deliberation network rather than a black-box generator. The framework operates through two interconnected stages: first, Phase I (Intuition-Guided Synthesis) addresses the ``anchoring'' problem by generating diverse expert intuitions and hierarchical plans to ground the reasoning, thereby preventing the model from drifting into hallucinated trajectories. Subsequently, Phase II (Adversarial Verification Loop) implements a recursive Skeptic-Researcher-Judge mechanism. By enforcing discrete verification gates and parametric probing, MAVEN forces the model to perform ``introspection'' and ``internal critique'' before narrative integration, effectively isolating parametric probing from narrative synthesis. Extensive evaluations on several datasets demonstrate that MAVEN not only achieves competitive accuracy against 10 state-of-the-art baselines but also significantly outperforms existing paradigms in reasoning rigor and justification quality, as evidenced by our proposed fine-grained metrics for causal depth and factual grounding.

In summary, our contributions are three-fold:

\begin{itemize}[nosep, leftmargin=1.5em]
    \item  Formulating structured reasoning as a decoupled process of deliberation, verification, and refinement, we introduce MAVEN, a novel multi-agent framework that operationalizes this principled decomposition via a blackboard architecture with adversarial verification loops.
    \item We implement a modular reasoning platform integrating role-specific agents and discrete verification gates, enabling iterative error correction and significantly enhancing the causal rigor and information density of generated explanations.

    \item Extensive experiments on multi-hop reasoning benchmarks demonstrate that MAVEN maintains competitive accuracy while achieving superior reasoning quality, yielding consistent gains in JCD, F\&C, C\&A and ARS metrics over state-of-the-art baselines.
\end{itemize}

\section{Related Work}
\textbf{From Prompting to Multi-Agent Deliberation.}
LLM reasoning has evolved from linear prompting~\citep{wei2022chain,zhou2022least,wang2023plan} to search-based~\citep{yao2023tree,besta2024graph} and autonomous structures~\citep{zhou2024self}. Drawing on the "Society of Mind"~\citep{minsky1988society}, multi-agent ecosystems now employ symmetric debate~\citep{chen2024reconcile,du2023improving}, asymmetric personas~\citep{chan2023chateval,wang2024mixture}, and collaborative frameworks spanning evolutionary to software-centric models~\citep{wynn2025talk,yuan2024evoagent,nascimento2023self,qian2023experimental,zhao2024electoral,wu2024spontaneous,qian2024chatdev,hong2024metagpt}. However, many systems still optimize for terminal agreement and may propagate early mistakes when consensus is reached without explicit verification gates. Echoing the view that correctness alone does not imply trustworthy reasoning~\citep{kadavath2022knowledge,zhao2024knowing}, MAVEN emphasizes a complete, verifiable logical chain. It adopts a blackboard architecture for decoupled orchestration, adversarial deliberation, and structural replanning, enabling agents to challenge intermediate assumptions rather than only reconcile final outputs. By extending feedback-driven refinement and step-wise validation~\citep{madaan2023selfrefine,shinn2023reflexion,lightman2024verify}, MAVEN prioritizes auditable, cross-agent accountability over static convergence.

\textbf{Factuality Verification and Evaluation.}
To mitigate LLM hallucinations~\citep{ji2023survey}, prior work explored confidence calibration~\citep{kadavath2022knowledge,zhang2024rtuning}, self-detection~\citep{zhao2024knowing}, and cross-examination~\citep{cohen2023lm}, alongside fine-grained evaluation rubrics~\citep{ye2024flask,pal2023medhalt}. While recent latent reasoners demonstrate strong capabilities~\citep{deepseek2025r1}, their opaque nature limits auditability and error attribution. Researchers have thus combined multi-agent deliberation~\citep{yue2023llmcascades,chen2023autoagents} with iterative verify-and-edit mechanisms~\citep{gao2023rarr,zhao2023verify,schick2023toolformer}. Although these loops and step-wise validations~\citep{madaan2023selfrefine,lightman2024verify} often improve outcomes, they can still converge without causal transparency. MAVEN addresses this gap by mandating cross-agent verification at each reasoning step; by grounding state transitions in verified evidence, it offers an interpretable alternative that synergizes with ReAct-style coupling~\citep{yao2023react} and logic-enhanced reasoning~\citep{pan2023logiclm}.

\section{Methodology}
\label{sec:methodology}

\begin{figure*}[t]
    \centering
    \includegraphics[width=\textwidth]{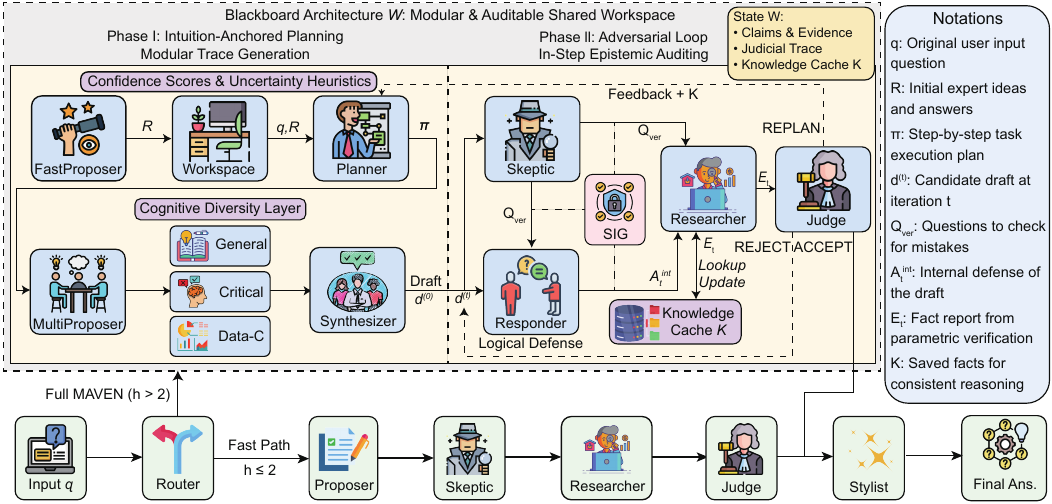} 
    \caption{Overview of MAVEN. MAVEN utilizes a blackboard architecture to decouple factual probing from synthesis. Following an Adaptive Router, Phase~I grounds hierarchical plans in Expert Intuitions to generate initial drafts. Phase~II then initiates an adversarial loop for forensic verification. Throughout the process, a persistent Knowledge Cache enables reasoning resets while preserving verified facts, effectively mitigating hallucination accumulation.}
    \label{fig:framework}
\end{figure*}

\subsection{Problem Formulation}
\label{subsec:formulation}

Given a complex query $q$ requiring multi-hop reasoning, our goal is to produce a final answer $a^*$ that is both factually accurate and logically coherent. We formalize this as an iterative refinement process over a shared workspace $\mathcal{W}$, coordinated by specialized agents through structured interventions.

At each discrete time step $t$, the workspace state is defined as:
\begin{equation}
    \mathcal{W}_t = \langle q, \mathbf{R}, \pi, d^{(t)}, \mathcal{A}_t^{\text{int}}, \mathcal{K}_t, \mathcal{V}_t \rangle
\end{equation}
where $q$ is the input query; $\mathbf{R} = \{r_1, \ldots, r_N\}$ denotes the set of $N$ expert intuitions; $\pi$ is the hierarchical reasoning plan; $d^{(t)}$ is the candidate draft at iteration $t$; $\mathcal{A}_t^{\text{int}}$ represents the internal justifications generated during deliberation; $\mathcal{K}_t$ is a persistent knowledge cache mapping verified claims to supporting evidence; and $\mathcal{V}_t$ is the judicial trace recording all deliberation decisions.

\subsection{Architecture Overview}
\label{subsec:architecture}

Figure~\ref{fig:framework} illustrates the overall architecture. MAVEN is built on the blackboard architecture paradigm, where the global workspace $\mathcal{W}$ functions as a ``cognitive ledger'' enabling state-aware interventions by specialized agents. Unlike linear pipelines that accumulate errors, MAVEN facilitates recursive feedback loops where reasoning states are dynamically updated and scrutinized. This centralized workspace manages cross-agent dependencies, allowing the system to reset reasoning paths when contradictions are detected. The architecture comprises two phases: (I) Intuition-Guided Synthesis, which anchors initial plans in expert knowledge, and (II) Adversarial Deliberation, which subjects drafts to iterative refinement through forensic auditing. By decoupling factual probing from synthesis, MAVEN maintains structural integrity even as query complexity scales.

To optimize computational efficiency, a Router agent estimates the reasoning complexity $h$---defined as the number of required factual verifications or logical inferences---and directs queries accordingly:
\[
\text{Path}(q)=
\textsc{Fast\_Path}\ \text{if } h\le\tau;\ 
\textsc{Full\_Maven}\ \text{if } h>\tau.
\]
where $\tau = 2$ is the complexity threshold. Simple queries bypass the full deliberation loop via the Fast Path, while complex queries activate the complete pipeline. A core design principle is the categorical separation of generation, critique, retrieval, and adjudication into isolated operational stages. This prevents ``hallucination drift'' (an agent's stochastic biases contaminate collective reasoning) by ensuring that no single agent controls both claim generation and verification.

\subsection{Phase I: Modular Intuition-Guided Synthesis}
\label{subsec:phase1}

A primary failure mode in agentic planning is ``unbounded drifting,'' where planners generate logically sound but factually ungrounded roadmaps. MAVEN addresses this through a three-stage process.

Stage 1: Expert Intuition Anchoring. A set of $N$ independent FastProposer agents generate intuitive responses $\mathbf{R} = \{r_1, \ldots, r_N\}$, each paired with an associated confidence score. These confidence scores and the variance across $\mathbf{R}$ serve as uncertainty heuristics: high consensus indicates straightforward tasks, while high divergence signals latent ambiguity requiring deeper verification.

Stage 2: Grounded Planning. The Planner agent transforms $q$ and $\mathbf{R}$ into a structured reasoning plan $\pi$. Critically, the Planner is designed to detect contradictions within $\mathbf{R}$ and incorporate explicit verification steps, ensuring that $\pi$ is anchored in identified uncertainties rather than assumed facts.

Stage 3: Divergent Perspective Synthesis. To mitigate premature cognitive closure  and ensure divergent perspective coverage, a MultiProposer agent generates candidate drafts through a Cognitive Diversity Layer encompassing three analytical lenses: (i) General, focusing on systematic drafting aligned with $\pi$; (ii) Critical, proactively interrogating potential risks; and (iii) Data-Centric, prioritizing quantitative and empirical evidence. Subsequently, the Synthesizer reconciles these perspectives—identifying consensus and resolving contradictions based on evidential strength—to integrate unique insights into a unified initial draft $d^{(0)}$.

\subsection{Phase II: Adversarial Deliberation with In-Step Epistemic Auditing}
\label{subsec:phase2}

The core of MAVEN lies in its recursive adversarial loop, where drafts undergo forensic auditing through four functionally decoupled agents: Skeptic, Responder, Researcher, and Judge.

Skeptic (Forensic Scrutiny).
The Skeptic performs a rigorous critique of the current draft $d^{(t)}$ along four investigative dimensions: (i) Fact Verification: probing numerical claims and statistics; (ii) Logical Coherence: identifying contradictions or non-sequiturs; (iii) Causal Scrutiny: questioning oversimplified cause-effect relationships; and (iv) Adversarial Questioning: challenging core assumptions. This produces a set of verification questions $\mathcal{Q}_{\text{ver}}$, each tagged with type and priority.

Responder (Internal Defense).
The Responder justifies $d^{(t)}$ against $\mathcal{Q}_{\text{ver}}$ using only parametric knowledge, without external retrieval. By forcing explicit articulation of the reasoning chain, this stage exposes latent assumptions and potential biases, producing internal justifications $\mathcal{A}_t^{\text{int}}$ that make the model's logical defense transparent and auditable.

Researcher (Parametric Probing).
The Researcher performs independent fact-checking by: (i) validating specific claims in $\mathcal{Q}_{\text{ver}}$ through parametric knowledge probing; (ii) contrasting verified facts against $\mathcal{A}_t^{\text{int}}$; and (iii) calibrating confidence in findings. This yields an evidence report $E_t$ and updates the persistent knowledge cache:
\begin{equation}
    \mathcal{K}_{t+1} = \mathcal{K}_t \cup \{(c, e) \mid c \in \mathcal{Q}_{\text{ver}},\; e = \textsc{Verify}(c)\}
\end{equation}
The decoupling of internal defense from independent parametric verification prevents propagation of plausible but incorrect reasoning.

Judge (Adjudication).
The Judge issues verdicts based on multi-dimensional evaluation across four criteria: factual accuracy, logical coherence, completeness (alignment with $\pi$), and neutrality. Based on the plan $\pi$, current draft $d^{(t)}$, and evidence report $E_t$, the Judge produces a verdict $\delta \in \{\textsc{Accept}, \textsc{Reject}, \textsc{Replan}\}$ that determines state transitions: Accept finalizes the draft for stylistic refinement; Reject triggers targeted revision producing $d^{(t+1)}$; Replan generates a new plan $\pi$ incorporating feedback while preserving $\mathcal{K}_t$.

Finalization.
Upon an Accept verdict, a Stylist agent performs linguistic refinement on the final draft, improving fluency and readability without altering the factual content or logical structure.

\subsection{Deliberation Pipeline and Robustness}
\label{subsec:pipeline}

The execution flow of MAVEN is summarized in Algorithm~\ref{alg:maven_main}. The framework follows a bounded iterative deliberation process that terminates either upon a Judge Accept verdict or when the maximum iteration budget $T_{\max}$ is reached. To ensure stable multi-agent execution, MAVEN integrates two robustness mechanisms. The Structural Integrity Gate (SIG) validates all inter-agent communications against predefined schemas before state transitions, employing a three-tier extraction strategy—direct JSON parsing, regex-based extraction upon failure, and Markdown code fence isolation as fallback—with schema violations triggering error traces for agent self-correction to ensure workspace consistency under stochastic LLM formatting variations. Additionally, the knowledge cache $\mathcal{K}$ persists across Replan cycles: when narrative resets occur, drafts and plans are discarded but verified facts are retained, enabling backend consistency check. While $\mathcal{K}$ prevents redundant probing, the Judge's textual feedback explicitly informs the Planner of failed claims to guide logical replanning.

\begin{algorithm}[t]
  \caption{MAVEN Deliberation Workflow}
  \label{alg:maven_main}
  \SetKwInput{KwIn}{Input}
  \SetKwInput{KwOut}{Output}
  \DontPrintSemicolon

  \KwIn{Query $q$, iteration limit $T_{\max}$}
  \KwOut{Final answer $a^*$}
  
  \lIf{$\textsc{Router}(q) = \text{FAST\_PATH}$}{\Return $\textsc{FastPath}(q)$}
  $R \gets \textsc{FastProposer}(q)$;\enspace $\pi \gets \textsc{Planner}(q, R)$;\enspace $d \gets \textsc{Synthesizer}(\textsc{MultiPropose}(q, \pi, R))$;\enspace $\mathcal{K} \gets \emptyset$\;
  
  \For{$t \gets 0$ \KwTo $T_{\max}-1$}{
      $\mathcal{Q} \gets \textsc{Skeptic}(d)$;\enspace $E, \mathcal{K} \gets \textsc{Researcher}(\mathcal{Q}, \textsc{Responder}(d, \mathcal{Q}), \mathcal{K})$\;
      $\delta, d', f \gets \textsc{Judge}(\pi, d, E)$ \tcp*{$f$ is reasoning feedback}
      \lIf{$\delta = \textsc{Accept}$}{\Return $\textsc{Stylist}(d)$}
      \lIf{$\delta = \textsc{Replan}$}{$\pi \gets \textsc{Planner}(q, f)$;\enspace $d \gets \textsc{Synthesizer}(\textsc{MultiPropose}(q, \pi, R))$}\lElse{$d \gets d'$}
  }
  \Return $\textsc{Stylist}(d)$\;
\end{algorithm}

\section{Experiments}
\label{sec:experiments}

\subsection{Experimental Settings}
\label{subsec:experimental_settings}

To comprehensively evaluate MAVEN's capability in logical deduction, hallucination mitigation, and factual robustness, we benchmark the framework across OpenBookQA (OBQA) \cite{mihaylov2018can}, TruthfulQA (TQA) \cite{lin2021truthfulqa}, HaluEval \cite{li2023halueval} and StrategyQA (SQA) \cite{geva2021did}. For each dataset, we randomly sample 300 instances from the official test set. We verify the statistical robustness of our comparisons using McNemar's test for accuracy and the Wilcoxon signed-rank test for quality scores, marking significant improvements ($p<0.05$) with $^\dagger$ in our result tables.

As shown in Table~\ref{tab:main_results}, we compare MAVEN against two baseline categories: agentic reasoning frameworks and frontier LLMs in zero-shot regimes. MAVEN uses DeepSeek-V3.2 as the backbone with deliberation constrained to $T_{\max}=3$ iterations and temperature 0.0. We adopt an LLM-as-a-Judge protocol assessing justification quality (1–100 scale) across four dimensions: Justification Plausibility and Causal Depth (JCD), Factual Accuracy and Contextualization (F\&C), Completeness and Comparative Analysis (C\&A), and Argument Richness and Structure (ARS). A blind human preference study additionally validates the automated evaluation protocol.

\subsection{Performance on Benchmarks}

Table~\ref{tab:main_results} summarizes answer accuracy (Acc) and four justification metrics (JCD, F\&C, C\&A, ARS) on OBQA, TQA, HaluEval, and SQA. MAVEN consistently delivers the strongest explanation quality across all datasets: it ranks best on JCD, C\&A, and ARS in every benchmark, and achieves the best F\&C on OBQA, HaluEval, and SQA, while remaining close to the top on TQA (90.90 vs.\ 91.37 for Gemini-3.1-Pro). Accuracy stays competitive—fourth on OBQA (96.33 vs.\ 98.00) and tied for third on HaluEval (98.33 vs.\ 99.00)—though it is lower on TQA and SQA, where Mixture-of-Agents (93.67) and Gemini-3.1-Pro (85.67) lead. Overall, MAVEN’s blackboard-based, verification-driven deliberation markedly improves causal depth, grounding, completeness, and structural rigor of justifications without materially sacrificing task performance.

To validate that our improvements are not artifacts of the GPT-4o judge, we additionally conduct a blind human preference study. We randomly sample 50 instances from each dataset and recruit 5 annotators, all of whom are PhD students. For each instance, annotators are shown five anonymized responses (MAVEN, Mixture-of-Agents, Self-Discover, ReConcile and GEMINI-3.1-PRO) in randomized order, and are asked to select the best and second-best response using the same four explanation-quality criteria as in our automatic evaluation (JCD, F\&C, C\&A, ARS). Human preferences are consistent with the automatic evaluation results, with MAVEN receiving the highest top-1 and top-2 selections across datasets.

\begin{table}[t]
\caption{Comparison on four benchmarks reporting Acc and four explanation-quality dimensions. MAVEN delivers consistent justification-quality gains while maintaining competitive accuracy. $\dagger$ indicates the corresponding method is significantly worse than MAVEN ($p<0.05$) using McNemar's test for Acc and Wilcoxon signed-rank test for quality dimensions.}
\label{tab:main_results}
\centering
\setlength{\tabcolsep}{2pt}

\begin{tabular}{l ccccc ccccc}
\toprule
& \multicolumn{5}{c}{\textbf{OPENBOOKQA}} & \multicolumn{5}{c}{\textbf{TRUTHFULQA}} \\
\cmidrule(lr){2-6} \cmidrule(lr){7-11}
\textbf{Method} & \textbf{ACC} & \textbf{JCD} & \textbf{F\&C} & \textbf{C\&A} & \textbf{ARS} & \textbf{ACC} & \textbf{JCD} & \textbf{F\&C} & \textbf{C\&A} & \textbf{ARS} \\
\midrule
LMVSLM & 94.00$^\dagger$ & 87.50$^\dagger$ & 90.07$^\dagger$ & 86.54$^\dagger$ & 90.05$^\dagger$ & 84.33$^\dagger$ & 87.57$^\dagger$ & 88.18$^\dagger$ & 84.99$^\dagger$ & 90.63$^\dagger$ \\
Mixture-of-Agents & 94.67 & 87.33$^\dagger$ & 89.93$^\dagger$ & 86.80$^\dagger$ & 89.40$^\dagger$ & \textbf{93.67} & 86.42$^\dagger$ & 90.32 & 86.64$^\dagger$ & 89.97$^\dagger$ \\
Multi-LLM Collab & 94.00$^\dagger$ & 88.70$^\dagger$ & 90.51$^\dagger$ & 87.87$^\dagger$ & 90.49$^\dagger$ & 87.00$^\dagger$ & 87.32$^\dagger$ & 89.07 & 86.18$^\dagger$ & 91.57$^\dagger$ \\
ReConcile & 95.33 & 84.47$^\dagger$ & 88.30$^\dagger$ & 80.57$^\dagger$ & 83.17$^\dagger$ & 79.67$^\dagger$ & 79.81$^\dagger$ & 83.37$^\dagger$ & 75.46$^\dagger$ & 84.35$^\dagger$ \\
Self-Detection & 95.00 & 86.60$^\dagger$ & 89.85$^\dagger$ & 86.34$^\dagger$ & 89.98$^\dagger$ & 87.00$^\dagger$ & 87.04$^\dagger$ & 88.62$^\dagger$ & 85.78$^\dagger$ & 90.43$^\dagger$ \\
Self-Discover & 96.67 & 89.47$^\dagger$ & 90.98$^\dagger$ & 88.21$^\dagger$ & 90.41$^\dagger$ & 87.33 & 86.91$^\dagger$ & 88.84$^\dagger$ & 85.02$^\dagger$ & 90.05$^\dagger$ \\

\midrule
DEEPSEEK-V3.2 & 95.67 & 82.91$^\dagger$ & 83.02$^\dagger$ & 80.96$^\dagger$ & 83.85$^\dagger$ & 85.67$^\dagger$ & 82.82$^\dagger$ & 87.52$^\dagger$ & 83.48$^\dagger$ & 85.97$^\dagger$ \\
DEEPSEEK-R1 & 97.33 & 78.73$^\dagger$ & 85.09$^\dagger$ & 67.92$^\dagger$ & 71.08$^\dagger$ & 90.00 & 84.62$^\dagger$ & 88.19$^\dagger$ & 84.04$^\dagger$ & 89.28$^\dagger$ \\
GEMINI-3.1-PRO & \textbf{98.00} & 86.33$^\dagger$ & 89.95$^\dagger$ & 84.42$^\dagger$ & 88.63$^\dagger$ & 92.33 & 89.43 & \textbf{91.37} & 88.11$^\dagger$ & 93.13$^\dagger$ \\
GPT-5.1 & 94.33$^\dagger$ & 83.22$^\dagger$ & 87.45$^\dagger$ & 79.32$^\dagger$ & 84.35$^\dagger$ & 90.00 & 87.53$^\dagger$ & 90.76 & 87.54$^\dagger$ & 92.30$^\dagger$ \\

\midrule
\textbf{MAVEN} & 96.33 & \textbf{92.92} & \textbf{93.77} & \textbf{89.92} & \textbf{96.10} & 90.00 & \textbf{90.77} & 90.90 & \textbf{89.37} & \textbf{96.38} \\

\midrule \midrule
& \multicolumn{5}{c}{\textbf{HALUEVAL}} & \multicolumn{5}{c}{\textbf{STRATEGYQA}} \\
\cmidrule(lr){2-6} \cmidrule(lr){7-11}
\textbf{Method} & \textbf{ACC} & \textbf{JCD} & \textbf{F\&C} & \textbf{C\&A} & \textbf{ARS} & \textbf{ACC} & \textbf{JCD} & \textbf{F\&C} & \textbf{C\&A} & \textbf{ARS} \\
\midrule
LMVSLM & 93.00$^\dagger$ & 88.37$^\dagger$ & 94.87$^\dagger$ & 87.14$^\dagger$ & 89.45$^\dagger$ & 82.00 & 89.67$^\dagger$ & 91.30$^\dagger$ & 84.64$^\dagger$ & 89.68$^\dagger$ \\
Mixture-of-Agents & 96.67 & 85.13$^\dagger$ & 93.90$^\dagger$ & 88.15$^\dagger$ & 87.30$^\dagger$ & 83.67 & 89.47$^\dagger$ & 91.77$^\dagger$ & 85.35$^\dagger$ & 89.82$^\dagger$ \\
Multi-LLM Collab & 91.00$^\dagger$ & 88.43$^\dagger$ & 93.15$^\dagger$ & 86.88$^\dagger$ & 89.85 & 81.67 & 90.12$^\dagger$ & 91.48$^\dagger$ & 86.08$^\dagger$ & 89.83$^\dagger$ \\
ReConcile & 96.67 & 85.44$^\dagger$ & 94.10$^\dagger$ & 87.75$^\dagger$ & 87.92$^\dagger$ & 78.00$^\dagger$ & 80.92$^\dagger$ & 86.23$^\dagger$ & 76.44$^\dagger$ & 86.37$^\dagger$ \\
Self-Detection & 98.67 & 89.20 & 97.34$^\dagger$ & 88.85$^\dagger$ & 89.83 & 82.33 & 89.02$^\dagger$ & 90.72$^\dagger$ & 84.55$^\dagger$ & 89.83$^\dagger$ \\
Self-Discover & 98.00 & 88.07$^\dagger$ & 97.52$^\dagger$ & 89.17$^\dagger$ & 87.15$^\dagger$ & 81.00 & 88.75$^\dagger$ & 90.78$^\dagger$ & 83.88$^\dagger$ & 89.52$^\dagger$ \\

\midrule
DEEPSEEK-V3.2 & 96.33 & 85.92$^\dagger$ & 94.18$^\dagger$ & 87.30$^\dagger$ & 83.85$^\dagger$ & 81.33 & 84.46$^\dagger$ & 85.48$^\dagger$ & 80.87$^\dagger$ & 86.49$^\dagger$ \\
DEEPSEEK-R1 & 98.00 & 87.58$^\dagger$ & 97.35$^\dagger$ & 89.25$^\dagger$ & 87.72 & 84.00 & 86.13$^\dagger$ & 89.92$^\dagger$ & 82.73$^\dagger$ & 87.77$^\dagger$ \\
GEMINI-3.1-PRO & 98.33 & 88.67$^\dagger$ & 97.95 & 89.18$^\dagger$ & 89.68 & \textbf{85.67} & 87.75$^\dagger$ & 90.13$^\dagger$ & 82.04$^\dagger$ & 87.17$^\dagger$ \\
GPT-5.1 & \textbf{99.00} & 85.67$^\dagger$ & 96.92$^\dagger$ & 89.30$^\dagger$ & 84.40$^\dagger$ & 84.67 & 86.78$^\dagger$ & 90.40$^\dagger$ & 82.77$^\dagger$ & 87.88$^\dagger$ \\

\midrule
\textbf{MAVEN} & 98.33 & \textbf{90.70} & \textbf{98.08} & \textbf{91.27} & \textbf{90.85} & 82.00 & \textbf{91.46} & \textbf{92.72} & \textbf{87.84} & \textbf{93.11} \\

\bottomrule
\end{tabular}
\end{table}

\subsection{Component Contribution}

To quantify the contribution of MAVEN’s core components, we conduct an ablation study on all four benchmarks (Table~\ref{tab:ablation}). Across datasets, the iterative deliberation loop is the dominant driver of performance: removing iterations ($T_{\max}=1$) causes the largest and most consistent degradation in both Acc and justification quality, with notable Acc drops on OBQA (96.33 to 93.33), TQA (90.00 to 87.00), and SQA (82.00 to 76.67), alongside substantial declines in JCD/C\&A/ARS (e.g., JCD 90.77 to 88.95 on TQA and 91.46 to 89.45 on SQA). Removing MultiProposer yields a milder but broad decrease across all metrics, suggesting that multi-perspective proposal diversity improves evidence coverage and reduces blind spots (e.g., OBQA JCD 92.92 to 90.26; HaluEval C\&A 91.27 to 90.83). Removing FastProposer also hurts robustness, particularly on Acc for OBQA/SQA/TQA (93.67/77.33/87.67), indicating that early high-confidence anchors help stabilize subsequent verification and refinement. Overall, the components act complementarily: iterations provide the core verification gains, while early anchoring and diverse proposals further improve coverage and justification structure.

\begin{table}[t]
\caption{Ablation results across four benchmarks. We report Acc and four explanation-quality dimensions (JCD, F\&C, C\&A, ARS). Each variant removes one component from the full framework; removing iterative deliberation ($T_{\max}=1$) yields the largest and most consistent drop across datasets.}
\label{tab:ablation}
\centering
\setlength{\tabcolsep}{4.5pt}
\begin{tabular}{l|ccccc|ccccc}
\toprule
\multirow{2}{*}{\textbf{Variant}} &
\multicolumn{5}{c|}{\textbf{OPENBOOKQA}} &
\multicolumn{5}{c}{\textbf{TRUTHFULQA}}  \\
\cmidrule(lr){2-6} \cmidrule(lr){7-11}
& \textbf{ACC} & \textbf{JCD} & \textbf{F\&C} & \textbf{C\&A} & \textbf{ARS}
& \textbf{ACC} & \textbf{JCD} & \textbf{F\&C} & \textbf{C\&A} & \textbf{ARS} \\
\midrule
Full Framework   & 96.33 & 92.92 & 93.77 & 89.92 & 96.10 & 90.00 & 90.77 & 90.90 & 89.37 & 96.38 \\
$-$ MultiProposer& 95.33 & 90.26 & 92.32 & 89.39 & 94.95 & 88.33 & 89.23 & 89.65 & 89.08 & 88.10 \\
$-$ Iterations   & 93.33 & 90.27 & 91.05 & 88.21 & 93.85 & 87.00 & 88.95 & 88.18 & 88.55 & 92.20 \\
$-$ FastProposer & 93.67 & 91.45 & 92.03 & 88.87 & 95.26 & 87.67 & 89.39 & 89.74 & 88.56 & 90.91 \\
\midrule
\multirow{2}{*}{\textbf{Variant}} &
\multicolumn{5}{c|}{\textbf{HALUEVAL}} &
\multicolumn{5}{c}{\textbf{STRATEGYQA}} \\
\cmidrule(lr){2-6} \cmidrule(lr){7-11}
& \textbf{ACC} & \textbf{JCD} & \textbf{F\&C} & \textbf{C\&A} & \textbf{ARS}
& \textbf{ACC} & \textbf{JCD} & \textbf{F\&C} & \textbf{C\&A} & \textbf{ARS} \\
\midrule
Full Framework   & 98.33 & 90.70 & 98.08 & 91.27 & 90.85 & 82.00 & 91.46 & 92.72 & 87.84 & 93.11 \\
$-$ MultiProposer& 97.67 & 90.37 & 97.77 & 90.83 & 90.42 & 80.67 & 90.33 & 91.01 & 87.32 & 92.06 \\
$-$ Iterations   & 97.33 & 89.99 & 97.43 & 90.49 & 90.06 & 76.67 & 89.45 & 90.75 & 86.03 & 90.10 \\
$-$ FastProposer & 97.67 & 90.12 & 97.67 & 90.63 & 90.28 & 77.33 & 90.62 & 91.50 & 86.68 & 91.11 \\
\bottomrule
\end{tabular}
\end{table}

\subsection{Qualitative Analysis and Case Study}
\label{subsec:qualitative_case_study}

\begin{table*}[t]
\centering
\caption{MAVEN Case Study. A demonstration from StrategyQA. Through adversarial deliberation, MAVEN upgrades its reasoning from superficial heuristics to rigorous ecological system analysis, successfully rectifying its initial causal gaps.}
\label{tab:case_study_abridged}
\resizebox{1.00\textwidth}{!}{
\begin{tabular}{@{}p{0.08\textwidth} p{0.92\textwidth}@{}}
\toprule
\multicolumn{2}{@{}l}{\textbf{User Query:} Are goldfish more difficult to care for than isopods? \quad \textbf{Ground Truth:} True (Yes)} \\
\midrule
\textbf{Initial Draft} & "...Goldfish require a large aquarium. The nitrogen cycle is mandatory and takes 4-8 weeks. Isopods require a physical humidity gradient which is simpler to create. Goldfish care is proactive, isopod care is reactive..." \\
\midrule
\textbf{\textsc{Skeptic}} & "Explain the precise biochemical mechanisms of the aquarium nitrogen cycle ($NH_3 \rightarrow NO_2^- \rightarrow NO_3^-$) versus isopod decomposition. Quantify the difference in microbial complexity and stability." \\

\textbf{\textsc{Resear- cher}} & "The aquarium nitrogen cycle is a linear, specialized chemoautotrophy dominated by a few genera. It is functionally fragile. The isopod enclosure relies on a web-like, generalized heterotrophy involving diverse kingdoms and is highly resilient due to functional redundancy." \\
\midrule
\textbf{\textsc{Judge}} & REJECT. Reasoning: The evidence report validates the technical claims but highlights that the draft's causal analysis of system stability is incomplete. The draft must be revised to integrate the precise biochemical vulnerabilities of aquatic vs. terrestrial microbial networks. \\
\midrule
\textbf{Revised Draft} & "...Goldfish live in a closed aquatic medium requiring a biological nitrogen cycle—a simplified, vulnerable metabolic pathway where specialized bacteria sequentially oxidize toxic ammonia. In contrast, isopods rely on a complex, resilient heterotrophic decomposition network providing a biological buffer. Therefore, goldfish are unequivocally more difficult due to the required precision of environmental control." \\
\bottomrule
\end{tabular}
}
\end{table*}

To demonstrate how MAVEN elevates reasoning quality, Table~\ref{tab:case_study_abridged} presents a representative trace from StrategyQA. While the initial draft correctly identified the ground truth (Yes), its justification relied on superficial heuristics. MAVEN’s adversarial loop immediately intervened: the Skeptic probed the underlying causal mechanisms, prompting the Researcher to anchor the comparison in empirical taxonomy (fragile aquatic chemoautotrophy vs. resilient terrestrial heterotrophy). Recognizing this epistemic gap, the Judge issued a Reject verdict, forcing the system to assimilate these scientific principles. The finalized draft successfully transformed a basic procedural comparison into a rigorous, biologically grounded analysis.

Broader qualitative analysis reveals that MAVEN's occasional underperformance in absolute accuracy relative to massive closed-source models stems from its conservative deliberation strategy. For straightforward queries, the Skeptic's hyper-vigilance can trigger over-correction or non-committal stances, marginally depressing rigid accuracy scores.
However, this reflects a deliberate architectural preference for cautious validation over unverified commitment. As shown in Table~\ref{tab:case_study_abridged}, this conservative scrutiny yields explicitly structured and auditable reasoning trajectories, preventing hallucination drift. In safety-critical domains (e.g., medicine or law), such interpretability and forensic verifiability make principled deliberation strictly preferable to opaque gains in raw accuracy.

\subsection{Iteration Count Sensitivity and System Stability}
\label{subsec:sensitivity}

\begin{figure}[t]
    \centering
    \includegraphics[width=1.0\textwidth]{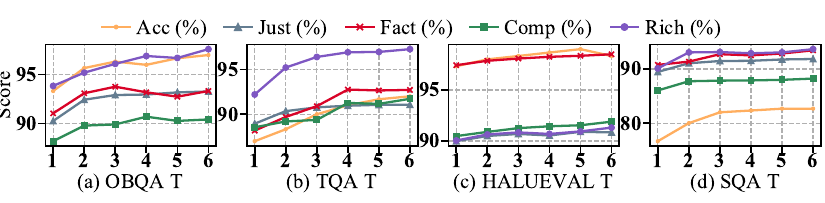} 
    \caption{Sensitivity analysis of $T_{max}$ (1--6) across four datasets. MAVEN exhibits asymptotic convergence. Performance leaps significantly from $T=1$ to $T=3$, after which the Judge agent's thresholding mechanism stabilizes the reasoning trajectory, preventing deliberative over-thinking.}
    \label{fig:sensitivity}
\end{figure}

To investigate the impact of deliberation depth on MAVEN's performance, we conduct a sensitivity analysis by varying $T_{max}$ from 1 to 6, where $T_{max}=1$ represents a baseline without adversarial deliberation. As shown in Figure~\ref{fig:sensitivity}, the steepest improvements occur between $T_{max}=1$ and $T_{max}=3$, validating that 1--2 rounds of forensic scrutiny effectively rectify initial hallucinations. Critically, MAVEN's curves do not exhibit deliberative over-thinking; instead, metrics plateau into stable asymptotes for $T \ge 4$. This robustness stems from the Judge Agent acting as a deterministic early-stopping gate once quality thresholds are met, and Epistemic Persistence of the knowledge cache preventing circular debates by anchoring verified facts. We recommend $T_{max}=3$ as the optimal configuration, balancing efficiency with deliberative rigor.

\subsection{Generalizability across Backbones}
\label{subsec:backbone_generalization}

\begin{table}[t]
\caption{Backbone generalization on four benchmarks. Results are reported as base model vs.\ base$+$MAVEN for two frontier models and two lightweight models. MAVEN yields consistent improvements in Acc and all explanation-quality dimensions (JCD, F\&C, C\&A, ARS), with particularly strong justification gains on lightweight and open-source backbones.}
\label{tab:backbone_results}
\centering
\setlength{\tabcolsep}{4pt}

\begin{tabular}{l ccccc ccccc}
\toprule
& \multicolumn{5}{c}{\textbf{OPENBOOKQA}} & \multicolumn{5}{c}{\textbf{TRUTHFULQA}} \\
\cmidrule(lr){2-6} \cmidrule(lr){7-11}
\textbf{Backbone / Method} & \textbf{ACC} & \textbf{JCD} & \textbf{F\&C} & \textbf{C\&A} & \textbf{ARS} & \textbf{ACC} & \textbf{JCD} & \textbf{F\&C} & \textbf{C\&A} & \textbf{ARS} \\
\midrule
DeepSeek-V3.2 Base         & 95.67 & 82.91 & 83.02 & 80.96 & 83.85 & 85.67 & 82.82 & 87.52 & 83.48 & 85.97 \\
\quad $+$ MAVEN       & 96.33 & 92.92 & 93.77 & 89.92 & 96.10 & 90.00 & 90.77 & 90.90 & 89.37 & 96.38 \\
GPT-5.1 Base          & 94.33 & 83.22 & 87.45 & 79.32 & 84.35 & 90.00 & 87.53 & 90.76 & 87.54 & 92.30 \\
\quad $+$ MAVEN       & 95.00 & 92.52 & 94.98 & 89.02 & 90.98 & 91.33 & 93.44 & 92.85 & 92.87 & 97.00 \\
\midrule
GPT-4o-mini Base      & 92.67 & 82.81 & 82.07 & 78.96 & 81.55 & 73.00 & 80.33 & 81.23 & 76.77 & 85.37 \\
\quad $+$ MAVEN       & 93.33 & 88.04 & 90.98 & 83.69 & 88.95 & 77.67 & 85.24 & 88.46 & 80.20 & 89.96 \\
Qwen3-8B Base         & 92.67 & 83.63 & 81.98 & 79.31 & 89.73 & 63.00 & 78.60 & 77.37 & 74.94 & 84.22 \\
\quad $+$ MAVEN       & 93.67 & 90.12 & 91.66 & 86.25 & 90.10 & 66.00 & 83.73 & 82.41 & 80.08 & 88.85 \\
\bottomrule
\end{tabular}

\begin{tabular}{l ccccc ccccc}
\toprule
& \multicolumn{5}{c}{\textbf{HALUEVAL}} & \multicolumn{5}{c}{\textbf{STRATEGYQA}} \\
\cmidrule(lr){2-6} \cmidrule(lr){7-11}
\textbf{Backbone / Method} & \textbf{ACC} & \textbf{JCD} & \textbf{F\&C} & \textbf{C\&A} & \textbf{ARS} & \textbf{ACC} & \textbf{JCD} & \textbf{F\&C} & \textbf{C\&A} & \textbf{ARS} \\
\midrule
DeepSeek-V3.2 Base         & 96.33 & 85.92 & 94.18 & 87.30 & 83.85 & 81.33 & 84.46 & 85.48 & 80.87 & 86.49 \\
\quad $+$ MAVEN       & 98.33 & 90.70 & 98.08 & 91.27 & 90.85 & 82.00 & 91.46 & 92.72 & 87.84 & 93.11 \\
GPT-5.1 Base          & 99.00 & 85.67 & 96.92 & 89.30 & 84.40 & 84.67 & 86.78 & 90.40 & 82.77 & 87.88 \\
\quad $+$ MAVEN       & 99.33 & 91.59 & 98.48 & 93.17 & 91.02 & 85.00 & 92.31 & 93.14 & 88.77 & 93.29 \\
\midrule
GPT-4o-mini Base      & 96.67 & 83.72 & 88.64 & 85.78 & 86.56 & 75.67 & 83.94 & 87.87 & 80.05 & 85.98 \\
\quad $+$ MAVEN       & 98.00 & 89.44 & 98.21 & 90.18 & 88.73 & 76.67 & 86.77 & 89.62 & 83.54 & 89.30 \\
Qwen3-8B Base         & 95.67 & 80.76 & 86.72 & 84.48 & 83.67 & 73.33 & 81.90 & 84.69 & 77.57 & 84.77 \\
\quad $+$ MAVEN       & 97.33 & 88.64 & 97.44 & 91.82 & 89.44 & 74.00 & 82.69 & 86.00 & 78.46 & 85.60 \\
\bottomrule
\end{tabular}
\end{table}

To assess MAVEN's model-agnostic nature, we evaluate it across diverse backbones including frontier models (GPT-5.1, DeepSeek-V3.2), lightweight variants (GPT-4o-mini), and open-source models (Qwen3-8B). As shown in Table~\ref{tab:backbone_results}, MAVEN consistently improves both accuracy and justification quality across all configurations and datasets. Frontier models achieve substantial gains: GPT-5.1+MAVEN attains the highest TQA scores in JCD (93.44) and ARS (97.00), while DeepSeek-V3.2+MAVEN shows pronounced improvements on OBQA, with JCD rising from 82.91 to 92.92 and ARS from 83.85 to 96.10, alongside similar gains on SQA where JCD increases from 84.46 to 91.46. The impact on lightweight models is equally notable—GPT-4o-mini+MAVEN achieves substantial F\&C improvements on HaluEval (from 88.64 to 98.21) and TQA (from 81.23 to 88.46), while Qwen3-8B+MAVEN yields consistent enhancements across all metrics on HaluEval, with JCD improving from 80.76 to 88.64 and C\&A from 84.48 to 91.82. Notably, justification-quality gains consistently exceed accuracy improvements across all backbones and datasets, indicating that MAVEN's explicit scaffolding systematically enhances structural rigor and logical coherence regardless of model scale.

\subsection{Computational Efficiency of Adaptive Routing}
\label{subsec:efficiency}

To evaluate the practical efficiency of MAVEN's Adaptive Complexity Router, we analyze the routing distribution and computational savings across all four benchmarks. The Router demonstrates clear sensitivity to task complexity, which is consistently reflected in both routing decisions and token usage: on StrategyQA, where implicit multi-hop reasoning is often required, 71.7\% of queries are routed to Full MAVEN and the average token consumption reaches 44.8k per query; in contrast, on HaluEval, whose queries demand less multi-hop reasoning, 96.3\% are routed through the Fast Path with a much lower average token usage of 10.6k tokens per query. Aggregating across all 1,200 test queries, Adaptive Routing reduces total wall-clock time by 43.5\% and token consumption by 43.9\%. These efficiency gains do not compromise reasoning quality, MAVEN achieves superior justification scores across all datasets despite routing approximately half of queries through the lightweight Fast Path, confirming that the Router accurately identifies queries where full deliberation provides marginal benefit.

\section{Conclusion}
\label{sec:conclusion}

We introduce MAVEN (Multi-Agent Verification-Elaboration Network with In-Step Epistemic Auditing), a stateful multi-agent framework that transforms large language models into explicit and auditable reasoners. By coupling expert intuition anchoring with an adversarial Skeptic–Researcher–Judge loop, MAVEN enforces causal rigor and systematic verification while remaining agnostic to the underlying backbone model. Empirical evaluations on OpenBookQA, TruthfulQA, HaluEval, and StrategyQA demonstrate consistent gains across causal depth, factual grounding, and consistency, outperforming both frontier LLMs and specialized agentic systems. The framework is particularly well suited for high-stakes domains where reliability and interpretability are critical. Current limitations include computational overhead from iterative multi-agent calls and verification bounded by parametric memory. Future work will integrate RAG pipelines and real-time web search to enable dynamic knowledge grounding beyond the model's training cutoff.

\bibliographystyle{unsrtnat}
\bibliography{references}

@article{wei2022chain,
  title={Chain-of-thought prompting elicits reasoning in large language models},
  author={Wei, Jason and Wang, Xuezhi and Schuurmans, Dale and Bosma, Maarten and Xia, Fei and Chi, Ed and Le, Quoc V and Zhou, Denny and others},
  journal={Advances in neural information processing systems},
  volume={35},
  pages={24824--24837},
  year={2022}
}

@article{yao2023tree,
  title={Tree of thoughts: Deliberate problem solving with large language models},
  author={Yao, Shunyu and Yu, Dian and Zhao, Jeffrey and Shafran, Izhak and Griffiths, Tom and Cao, Yuan and Narasimhan, Karthik},
  journal={Advances in neural information processing systems},
  volume={36},
  pages={11809--11822},
  year={2023}
}

@article{wang2024mixture,
  title={Mixture-of-agents enhances large language model capabilities},
  author={Wang, Junlin and Wang, Jue and Athiwaratkun, Ben and Zhang, Ce and Zou, James},
  journal={arXiv preprint arXiv:2406.04692},
  year={2024}
}

@inproceedings{chen2024reconcile,
  title={Reconcile: Round-table conference improves reasoning via consensus among diverse llms},
  author={Chen, Justin and Saha, Swarnadeep and Bansal, Mohit},
  booktitle={Proceedings of the 62nd Annual Meeting of the Association for Computational Linguistics (Volume 1: Long Papers)},
  pages={7066--7085},
  year={2024}
}

@article{zhou2024self,
  title={Self-discover: Large language models self-compose reasoning structures},
  author={Zhou, Pei and Pujara, Jay and Ren, Xiang and Chen, Xinyun and Cheng, Heng-Tze and Le, Quoc V and Chi, Ed and Zhou, Denny and Mishra, Swaroop and Zheng, Huaixiu Steven},
  journal={Advances in Neural Information Processing Systems},
  volume={37},
  pages={126032--126058},
  year={2024}
}

@inproceedings{zhao2024knowing,
  title={Knowing what llms do not know: A simple yet effective self-detection method},
  author={Zhao, Yukun and Yan, Lingyong and Sun, Weiwei and Xing, Guoliang and Meng, Chong and Wang, Shuaiqiang and Cheng, Zhicong and Ren, Zhaochun and Yin, Dawei},
  booktitle={Proceedings of the 2024 conference of the north American chapter of the Association for Computational Linguistics: Human language technologies (Volume 1: Long Papers)},
  pages={7051--7063},
  year={2024}
}

@article{cohen2023lm,
  title={Lm vs lm: Detecting factual errors via cross examination},
  author={Cohen, Roi and Hamri, May and Geva, Mor and Globerson, Amir},
  journal={arXiv preprint arXiv:2305.13281},
  year={2023}
}

@inproceedings{feng2024don,
title={Don't Hallucinate, Abstain: Identifying LLM Knowledge Gaps via Multi-LLM Collaboration},
author={Feng, Shangbin and Shi, Weijia and Wang, Yike and others},
booktitle={Proceedings of the 62nd Annual Meeting of the ACL},
pages={14664--14690},
year={2024}
}

@article{ji2023survey,
  title={Survey of hallucination in natural language generation},
  author={Ji, Ziwei and Lee, Nayeon and Frieske, Rita and Yu, Tiezheng and Su, Dan and Xu, Yan and Ishii, Etsuko and Bang, Ye Jin and Madotto, Andrea and Fung, Pascale},
  journal={ACM computing surveys},
  volume={55},
  number={12},
  pages={1--38},
  year={2023},
  publisher={ACM New York, NY}
}

@book{minsky1988society,
  title = {Society of Mind},
  author = {Minsky, Marvin},
  publisher = {Simon and Schuster},
  year = {1988}
}

@article{nii1986blackboard,
  title={The blackboard model of problem solving and the evolution of blackboard architectures},
  author={Nii, H Penny},
  journal={AI magazine},
  volume={7},
  number={2},
  pages={38--38},
  year={1986}
}

@inproceedings{besta2024graph,
  title={Graph of thoughts: Solving elaborate problems with large language models},
  author={Besta, Maciej and Blach, Nils and Kubicek, Ales and Gerstenberger, Robert and Podstawski, Michal and Gianinazzi, Lukas and Gajda, Joanna and Lehmann, Tomasz and Niewiadomski, Hubert and Nyczyk, Piotr and others},
  booktitle={Proceedings of the AAAI conference on artificial intelligence},
  volume={38},
  number={16},
  pages={17682--17690},
  year={2024}
}

@inproceedings{du2023improving,
  title={Improving factuality and reasoning in language models through multiagent debate},
  author={Du, Yilun and Li, Shuang and Torralba, Antonio and Tenenbaum, Joshua B and Mordatch, Igor},
  booktitle={Forty-first International Conference on Machine Learning},
  year={2023}
}

@article{chan2023chateval,
  title={Chateval: Towards better llm-based evaluators through multi-agent debate},
  author={Chan, Chi-Min and Chen, Weize and Su, Yusheng and Yu, Jianxuan and Xue, Wei and Zhang, Shanghang and Fu, Jie and Liu, Zhiyuan},
  journal={arXiv preprint arXiv:2308.07201},
  year={2023}
}

@article{li2023camel,
  title={Camel: Communicative agents for" mind" exploration of large language model society},
  author={Li, Guohao and Hammoud, Hasan and Itani, Hani and Khizbullin, Dmitrii and Ghanem, Bernard},
  journal={Advances in Neural Information Processing Systems},
  volume={36},
  pages={51991--52008},
  year={2023}
}

@article{wang2023plan,
  title={Plan-and-solve prompting: Improving zero-shot chain-of-thought reasoning by large language models},
  author={Wang, Lei and Xu, Wanyu and Lan, Yihuai and Hu, Zhiqiang and Lan, Yunshi and Lee, Roy Ka-Wei and Lim, Ee-Peng},
  journal={arXiv preprint arXiv:2305.04091},
  year={2023}
}

@article{zhou2022least,
  title={Least-to-most prompting enables complex reasoning in large language models},
  author={Zhou, Denny and Sch{\"a}rli, Nathanael and Hou, Le and Wei, Jason and Scales, Nathan and Wang, Xuezhi and Schuurmans, Dale and Cui, Claire and Bousquet, Olivier and Le, Quoc and others},
  journal={arXiv preprint arXiv:2205.10625},
  year={2022}
}

@article{huang2025survey,
  title={A survey on hallucination in large language models: Principles, taxonomy, challenges, and open questions},
  author={Huang, Lei and Yu, Weijiang and Ma, Weitao and Zhong, Weihong and Feng, Zhangyin and Wang, Haotian and Chen, Qianglong and Peng, Weihua and Feng, Xiaocheng and Qin, Bing and others},
  journal={ACM Transactions on Information Systems},
  volume={43},
  number={2},
  pages={1--55},
  year={2025},
  publisher={ACM New York, NY}
}

@article{kadavath2022knowledge,
title={Language models (mostly) know what they know},
author={Kadavath, Saurav and Conerly, Tom and Askell, Amanda and others},
journal={arXiv preprint arXiv:2207.05221},
year={2022}
}

@inproceedings{ye2024flask,
title={{FLASK}: Fine-grained Language Model Evaluation based on Alignment Skill Sets},
author={Seonghyeon Ye and Doyoung Kim and Sungdong Kim and Hyeonbin Hwang and Seungone Kim and Yongrae Jo and James Thorne and Juho Kim and Minjoon Seo},
booktitle={The Twelfth International Conference on Learning Representations},
year={2024},
url={https://openreview.net/forum?id=CYmF38ysDa}
}

@inproceedings{zhang2024rtuning,
    title = "{R}-Tuning: Instructing Large Language Models to Say `{I} Don{'}t Know'",
    author = "Zhang, Hanning  and
      Diao, Shizhe  and
      Lin, Yong  and
      Fung, Yi  and
      Lian, Qing  and
      Wang, Xingyao  and
      Chen, Yangyi  and
      Ji, Heng  and
      Zhang, Tong",
    editor = "Duh, Kevin  and
      Gomez, Helena  and
      Bethard, Steven",
    booktitle = "Proceedings of the 2024 Conference of the North American Chapter of the Association for Computational Linguistics: Human Language Technologies (Volume 1: Long Papers)",
    month = jun,
    year = "2024",
    address = "Mexico City, Mexico",
    publisher = "Association for Computational Linguistics",
    url = "https://aclanthology.org/2024.naacl-long.394/",
    doi = "10.18653/v1/2024.naacl-long.394",
    pages = "7113--7139"
}

@article{deepseek2025r1,
title={DeepSeek-R1: Incentivizing Reasoning Capability in LLMs via Reinforcement Learning},
author={DeepSeek-AI},
journal={arXiv preprint arXiv:2501.12948},
year={2025}
}

@inproceedings{pal2023medhalt,
    title = "{M}ed-{HALT}: Medical Domain Hallucination Test for Large Language Models",
    author = "Pal, Ankit  and
      Umapathi, Logesh Kumar  and
      Sankarasubbu, Malaikannan",
    editor = "Jiang, Jing  and
      Reitter, David  and
      Deng, Shumin",
    booktitle = "Proceedings of the 27th Conference on Computational Natural Language Learning (CoNLL)",
    month = dec,
    year = "2023",
    address = "Singapore",
    publisher = "Association for Computational Linguistics",
    url = "https://aclanthology.org/2023.conll-1.21/",
    doi = "10.18653/v1/2023.conll-1.21",
    pages = "314--334"
}

@techreport{openai2023gpt4,
  author       = {{OpenAI}},
  title        = {GPT-4 Technical Report},
  institution  = {OpenAI},
  year         = {2023},
  month        = mar,
  note         = {Version 6 (last revised 4 March 2024)},
  url          = {https://cdn.openai.com/papers/gpt-4.pdf},
  eprint       = {2303.08774},
  archivePrefix= {arXiv},
  primaryClass = {cs.CL}
}

@techreport{deepseek2024v3,
  author       = {{DeepSeek-AI}},
  title        = {DeepSeek-V3 Technical Report},
  institution  = {DeepSeek-AI},
  year         = {2024},
  month        = dec,
  note         = {arXiv:2412.19437 [cs.CL], last revised February 2025 (v2)},
  url          = {https://arxiv.org/pdf/2412.19437},
  eprint       = {2412.19437},
  archivePrefix= {arXiv},
  primaryClass = {cs.CL}
}

@inproceedings{wang2022self,
  author       = {Wang, Xuezhi and Wei, Jason and Schuurmans, Dale and Le, Quoc and Chi, Ed H. and Narang, Sharan and Chowdhery, Aakanksha and Zhou, Denny},
  title        = {Self-Consistency Improves Chain of Thought Reasoning in Language Models},
  booktitle    = {The Eleventh International Conference on Learning Representations},
  year         = {2023},
  url          = {https://openreview.net/forum?id=1PL1NIMMrw},
  note         = {ICLR 2023 poster},
  eprint       = {2203.11171},
  archivePrefix= {arXiv},
  primaryClass = {cs.CL}
}

@inproceedings{mihaylov2018can,
  title     = {Can a Suit of Armor Conduct Electricity? {A} New Dataset for Open Book Question Answering},
  author    = {Mihaylov, Todor and Clark, Peter and Khot, Tushar and Sabharwal, Ashish},
  booktitle = {Proceedings of the 2018 Conference on Empirical Methods in Natural Language Processing},
  pages     = {2381--2391},
  year      = {2018},
  month     = oct,
  address   = {Brussels, Belgium},
  publisher = {Association for Computational Linguistics},
  doi       = {10.18653/v1/D18-1260},
  url       = {https://aclanthology.org/D18-1260},
  eprint    = {1809.02789},
  archivePrefix = {arXiv},
  primaryClass = {cs.CL}
}

@inproceedings{lin2021truthfulqa,
  author       = {Lin, Stephanie and Hilton, Jacob and Evans, Owain},
  title        = {Truthful{QA}: Measuring How Models Mimic Human Falsehoods},
  booktitle    = {Proceedings of the 60th Annual Meeting of the Association for Computational Linguistics (Volume 1: Long Papers)},
  pages        = {3214--3252},
  year         = {2022},
  month        = may,
  address      = {Dublin, Ireland},
  publisher    = {Association for Computational Linguistics},
  doi          = {10.18653/v1/2022.acl-long.229},
  url          = {https://aclanthology.org/2022.acl-long.229},
  eprint       = {2109.07958},
  archivePrefix= {arXiv},
  primaryClass = {cs.CL}
}

@inproceedings{yuan2024evoagent,
  author       = {Yuan, Siyu and Song, Kaitao and Chen, Jiangjie and Tan, Xu and Li, Dongsheng and Yang, Deqing},
  title        = {{EvoAgent}: Towards Automatic Multi-Agent Generation via Evolutionary Algorithms},
  booktitle    = {Proceedings of the 2025 Conference of the Nations of the Americas Chapter of the Association for Computational Linguistics: Human Language Technologies (Volume 1: Long Papers)},
  pages        = {6192--6217},
  year         = {2025},
  month        = apr,
  address      = {Albuquerque, New Mexico},
  publisher    = {Association for Computational Linguistics},
  doi          = {10.18653/v1/2025.naacl-long.315},
  url          = {https://aclanthology.org/2025.naacl-long.315},
  eprint       = {2406.14228},
  archivePrefix= {arXiv},
  primaryClass = {cs.AI}
}

@article{wynn2025talk,
  author       = {Wynn, Andrea and Satija, Harsh and Hadfield, Gillian},
  title        = {Talk Isn't Always Cheap: Understanding Failure Modes in Multi-Agent Debate},
  journal      = {arXiv preprint arXiv:2509.05396},
  year         = {2025},
  month        = sep,
  note         = {v2 (13 Oct 2025); Accepted to ICML 2025 Multi-Agent Systems Workshop},
  url          = {https://arxiv.org/abs/2509.05396},
  eprint       = {2509.05396},
  archivePrefix= {arXiv},
  primaryClass = {cs.CL}
}

@inproceedings{nascimento2023self,
  author       = {Nascimento, Nathalia and Alencar, Paulo and Cowan, Donald},
  title        = {Self-Adaptive Large Language Model ({LLM})-Based Multiagent Systems},
  booktitle    = {2023 IEEE International Conference on Autonomic Computing and Self-Organizing Systems Companion (ACSOS-C)},
  pages        = {104--109},
  year         = {2023},
  month        = sep,
  address      = {Toronto, ON, Canada},
  publisher    = {IEEE},
  doi          = {10.1109/ACSOS-C58168.2023.00048},
  url          = {https://ieeexplore.ieee.org/document/10336211},
  eprint       = {2307.06187},
  archivePrefix= {arXiv},
  primaryClass = {cs.MA}
}

@inproceedings{chen2023autoagents,
  author       = {Chen, Guangyao and Dong, Siwei and Shu, Yu and Zhang, Ge and Sesay, Jaward and Karlsson, B{\"o}rje F. and Fu, Jie and Shi, Yemin},
  title        = {{AutoAgents}: A Framework for Automatic Agent Generation},
  booktitle    = {Proceedings of the Thirty-Third International Joint Conference on Artificial Intelligence (IJCAI)},
  pages        = {22--30},
  year         = {2024},
  month        = aug,
  address      = {Jeju Island, South Korea},
  publisher    = {ijcai.org},
  doi          = {10.24963/ijcai.2024/3},
  url          = {https://www.ijcai.org/proceedings/2024/0003},
  eprint       = {2309.17288},
  archivePrefix= {arXiv},
  primaryClass = {cs.AI}
}

@inproceedings{qian2023experimental,
  author       = {Qian, Chen and Dang, Yufan and Li, Jiahao and Liu, Wei and Xie, Zihao and Wang, Yifei and Chen, Weize and Yang, Cheng and Cong, Xin and Che, Xiaoyin and Liu, Zhiyuan and Sun, Maosong},
  title        = {Experiential Co-Learning of Software-Developing Agents},
  booktitle    = {Proceedings of the 62nd Annual Meeting of the Association for Computational Linguistics (Volume 1: Long Papers)},
  pages        = {5628--5640},
  year         = {2024},
  month        = aug,
  address      = {Bangkok, Thailand},
  publisher    = {Association for Computational Linguistics},
  doi          = {10.18653/v1/2024.acl-long.305},
  url          = {https://aclanthology.org/2024.acl-long.305},
  eprint       = {2312.17025},
  archivePrefix= {arXiv},
  primaryClass = {cs.CL}
}

@inproceedings{zhao2024electoral,
  author       = {Zhao, Xiutian and Wang, Ke and Peng, Wei},
  title        = {An Electoral Approach to Diversify {LLM}-based Multi-Agent Collective Decision-Making},
  booktitle    = {Proceedings of the 2024 Conference on Empirical Methods in Natural Language Processing},
  pages        = {2712--2727},
  year         = {2024},
  month        = nov,
  address      = {Miami, Florida, USA},
  publisher    = {Association for Computational Linguistics},
  doi          = {10.18653/v1/2024.emnlp-main.158},
  url          = {https://aclanthology.org/2024.emnlp-main.158},
  eprint       = {2410.15168},
  archivePrefix= {arXiv},
  primaryClass = {cs.CL}
}

@inproceedings{yue2023llmcascades,
  author       = {Yue, Murong and Zhao, Jie and Zhang, Min and Du, Liang and Yao, Ziyu},
  title        = {Large Language Model Cascades with Mixture of Thoughts Representations for Cost-efficient Reasoning},
  booktitle    = {The Twelfth International Conference on Learning Representations},
  year         = {2024},
  month        = may,
  url          = {https://openreview.net/forum?id=6okaSfANzh},
  note         = {ICLR 2024 poster},
  eprint       = {2310.03094},
  archivePrefix= {arXiv},
  primaryClass = {cs.CL}
}

@inproceedings{wu2024spontaneous,
  author       = {Wu, Zengqing and Peng, Run and Zheng, Shuyuan and Liu, Qianying and Han, Xu and Kwon, Brian I. and Onizuka, Makoto and Tang, Shaojie and Xiao, Chuan},
  title        = {Shall We Team Up: Exploring Spontaneous Cooperation of Competing {LLM} Agents},
  booktitle    = {Findings of the Association for Computational Linguistics: EMNLP 2024},
  pages        = {5163--5186},
  year         = {2024},
  month        = nov,
  address      = {Miami, Florida, USA},
  publisher    = {Association for Computational Linguistics},
  doi          = {10.18653/v1/2024.findings-emnlp.297},
  url          = {https://aclanthology.org/2024.findings-emnlp.297},
  eprint       = {2402.12327},
  archivePrefix= {arXiv},
  primaryClass = {cs.CL}
}

@article{madaan2023selfrefine,
  title     = {Self-Refine: Iterative Refinement with Self-Feedback},
  author    = {Madaan, Aman and Tandon, Niket and Gupta, Prakhar and Hallinan, Skyler and Gao, Luyu and Wiegreffe, Sarah and Alon, Uri and Dziri, Nouha and Prabhumoye, Shrimai and Yang, Yiming and Gupta, Shashank and Majumder, Bodhisattwa Prasad and Hermann, Katherine and Welleck, Sean and Yazdanbakhsh, Amir and Clark, Peter},
  journal   = {Advances in Neural Information Processing Systems},
  volume    = {36},
  pages     = {46534--46562},
  year      = {2023},
  note      = {NeurIPS 2023},
  url       = {https://proceedings.neurips.cc/paper_files/paper/2023/hash/91edff07232fb1b55a505a9e9f6c0ff3-Abstract-Conference.html},
  eprint    = {2303.17651},
  archivePrefix = {arXiv},
  primaryClass = {cs.CL}
}

@inproceedings{yao2023react,
  author       = {Yao, Shunyu and Zhao, Jeffrey and Yu, Dian and Du, Nan and Shafran, Izhak and Narasimhan, Karthik and Cao, Yuan},
  title        = {{ReAct}: Synergizing Reasoning and Acting in Language Models},
  booktitle    = {The Eleventh International Conference on Learning Representations},
  year         = {2023},
  month        = may,
  url          = {https://openreview.net/forum?id=WE_vluYUL-X},
  note         = {ICLR 2023 (notable top 5\%)},
  eprint       = {2210.03629},
  archivePrefix= {arXiv},
  primaryClass = {cs.CL}
}

@inproceedings{hong2024metagpt,
  author       = {Hong, Sirui and Zhuge, Mingchen and Chen, Jiaqi and Zheng, Xiawu and Cheng, Yuheng and Zhang, Ceyao and Wang, Jinlin and Wang, Zili and Yau, Steven Ka Shing and Lin, Zijuan and Zhou, Liyang and Ran, Chenyu and Xiao, Lingfeng and Wu, Chenglin and Schmidhuber, J{\"u}rgen},
  title        = {{Meta{GPT}}: Meta Programming for A Multi-Agent Collaborative Framework},
  booktitle    = {The Twelfth International Conference on Learning Representations},
  year         = {2024},
  month        = may,
  url          = {https://openreview.net/forum?id=VtmBAGCN7o},
  note         = {ICLR 2024 oral (top 1.2\%)},
  eprint       = {2308.00352},
  archivePrefix= {arXiv},
  primaryClass = {cs.AI}
}

@article{shinn2023reflexion,
  title     = {Reflexion: Language Agents with Verbal Reinforcement Learning},
  author    = {Shinn, Noah and Cassano, Federico and Labash, Beck and Gopinath, Ashwin and Narasimhan, Karthik and Yao, Shunyu},
  journal   = {Advances in Neural Information Processing Systems},
  volume    = {36},
  year      = {2023},
  note      = {NeurIPS 2023},
  url       = {https://proceedings.neurips.cc/paper_files/paper/2023/hash/1b44b878bb782e6954cd888628510e90-Abstract-Conference.html},
  eprint    = {2303.11366},
  archivePrefix = {arXiv},
  primaryClass = {cs.AI}
}

@inproceedings{gao2023rarr,
  author       = {Gao, Luyu and Dai, Zhuyun and Pasupat, Panupong and Chen, Anthony and Chaganty, Arun Tejasvi and Fan, Yicheng and Zhao, Vincent and Lao, Ni and Lee, Hongrae and Juan, Da-Cheng and Guu, Kelvin and Callison-Burch, Chris},
  title        = {{RARR}: Researching and Revising What Language Models Say, Using Language Models},
  booktitle    = {Proceedings of the 61st Annual Meeting of the Association for Computational Linguistics (Volume 1: Long Papers)},
  pages        = {16477--16508},
  year         = {2023},
  month        = jul,
  address      = {Toronto, Canada},
  publisher    = {Association for Computational Linguistics},
  doi          = {10.18653/v1/2023.acl-long.910},
  url          = {https://aclanthology.org/2023.acl-long.910},
  eprint       = {2210.08726},
  archivePrefix= {arXiv},
  primaryClass = {cs.CL}
}

@inproceedings{zhao2023verify,
  author       = {Zhao, Ruochen and Li, Xingxuan and Joty, Shafiq and Qin, Chengwei and Bing, Lidong},
  title        = {Verify-and-Edit: A Knowledge-Enhanced Chain-of-Thought Framework},
  booktitle    = {Proceedings of the 61st Annual Meeting of the Association for Computational Linguistics (Volume 1: Long Papers)},
  pages        = {5823--5840},
  year         = {2023},
  month        = jul,
  address      = {Toronto, Canada},
  publisher    = {Association for Computational Linguistics},
  doi          = {10.18653/v1/2023.acl-long.320},
  url          = {https://aclanthology.org/2023.acl-long.320},
  eprint       = {2305.03268},
  archivePrefix= {arXiv},
  primaryClass = {cs.CL}
}

@inproceedings{pan2023logiclm,
  author       = {Pan, Liangming and Albalak, Alon and Wang, Xinyi and Wang, William Yang},
  title        = {{Logic-LM}: Empowering Large Language Models with Symbolic Solvers for Faithful Logical Reasoning},
  booktitle    = {Findings of the Association for Computational Linguistics: EMNLP 2023},
  pages        = {3806--3824},
  year         = {2023},
  month        = dec,
  address      = {Singapore},
  publisher    = {Association for Computational Linguistics},
  doi          = {10.18653/v1/2023.findings-emnlp.248},
  url          = {https://aclanthology.org/2023.findings-emnlp.248},
  eprint       = {2305.12295},
  archivePrefix= {arXiv},
  primaryClass = {cs.CL}
}

@inproceedings{schick2023toolformer,
  author       = {Schick, Timo and Dwivedi-Yu, Jane and Dess{\`i}, Roberto and Raileanu, Roberta and Lomeli, Maria and Hambro, Eric and Zettlemoyer, Luke and Cancedda, Nicola and Scialom, Thomas},
  title        = {{Toolformer}: Language Models Can Teach Themselves to Use Tools},
  booktitle    = {Advances in Neural Information Processing Systems},
  volume       = {36},
  year         = {2023},
  note         = {NeurIPS 2023 oral},
  url          = {https://openreview.net/forum?id=Yacmpz84TH},
  eprint       = {2302.04761},
  archivePrefix= {arXiv},
  primaryClass = {cs.CL}
}

@inproceedings{qian2024chatdev,
  author       = {Qian, Chen and Liu, Wei and Liu, Hongzhang and Chen, Nuo and Dang, Yufan and Li, Jiahao and Yang, Cheng and Chen, Weize and Su, Yusheng and Cong, Xin and Xu, Juyuan and Li, Dahai and Liu, Zhiyuan and Sun, Maosong},
  title        = {{ChatDev}: Communicative Agents for Software Development},
  booktitle    = {Proceedings of the 62nd Annual Meeting of the Association for Computational Linguistics (Volume 1: Long Papers)},
  pages        = {15174--15186},
  year         = {2024},
  month        = aug,
  address      = {Bangkok, Thailand},
  publisher    = {Association for Computational Linguistics},
  doi          = {10.18653/v1/2024.acl-long.810},
  url          = {https://aclanthology.org/2024.acl-long.810},
  eprint       = {2307.07924},
  archivePrefix= {arXiv},
  primaryClass = {cs.SE}
}

@inproceedings{lightman2024verify,
  author       = {Lightman, Hunter and Kosaraju, Vineet and Burda, Yura and Edwards, Harri and Baker, Bowen and Lee, Teddy and Leike, Jan and Schulman, John and Sutskever, Ilya and Cobbe, Karl},
  title        = {Let's Verify Step by Step},
  booktitle    = {The Twelfth International Conference on Learning Representations},
  year         = {2024},
  month        = may,
  url          = {https://openreview.net/forum?id=v8L0pN6EOi},
  note         = {ICLR 2024 poster},
  eprint       = {2305.20050},
  archivePrefix= {arXiv},
  primaryClass = {cs.LG}
}

@inproceedings{li2023halueval,
  title={Halueval: A large-scale hallucination evaluation benchmark for large language models},
  author={Li, Junyi and Cheng, Xiaoxue and Zhao, Xin and Nie, Jian-Yun and Wen, Ji-Rong},
  booktitle={Proceedings of the 2023 conference on empirical methods in natural language processing},
  pages={6449--6464},
  year={2023}
}

@article{geva2021did,
  title={Did aristotle use a laptop? a question answering benchmark with implicit reasoning strategies},
  author={Geva, Mor and Khashabi, Daniel and Segal, Elad and Khot, Tushar and Roth, Dan and Berant, Jonathan},
  journal={Transactions of the Association for Computational Linguistics},
  volume={9},
  pages={346--361},
  year={2021},
  publisher={MIT Press One Rogers Street, Cambridge, MA 02142-1209, USA journals-info~…}
}

@article{qian2024scaling,
  title={Scaling large language model-based multi-agent collaboration},
  author={Qian, Chen and Xie, Zihao and Wang, Yifei and Liu, Wei and Zhu, Kunlun and Xia, Hanchen and Dang, Yufan and Du, Zhuoyun and Chen, Weize and Yang, Cheng and others},
  journal={arXiv preprint arXiv:2406.07155},
  year={2024}
}

@inproceedings{liu2025breaking,
  title={Breaking mental set to improve reasoning through diverse multi-agent debate},
  author={Liu, Yexiang and Cao, Jie and Li, Zekun and He, Ran and Tan, Tieniu},
  booktitle={The Thirteenth International Conference on Learning Representations},
  year={2025}
}

@article{dang2025multi,
  title={Multi-agent collaboration via evolving orchestration},
  author={Dang, Yufan and Qian, Chen and Luo, Xueheng and Fan, Jingru and Xie, Zihao and Shi, Ruijie and Chen, Weize and Yang, Cheng and Che, Xiaoyin and Tian, Ye and others},
  journal={arXiv preprint arXiv:2505.19591},
  year={2025}
}

@inproceedings{li2025advancing,
  title={Advancing collaborative debates with role differentiation through multi-agent reinforcement learning},
  author={Li, Haoran and Su, Ziyi and Xue, Yun and Tian, Zhiliang and Song, Yiping and Huang, Minlie},
  booktitle={Proceedings of the 63rd Annual Meeting of the Association for Computational Linguistics (Volume 1: Long Papers)},
  pages={22655--22666},
  year={2025}
}

@article{li2025reasoning,
  title={Reasoning models hallucinate more: Factuality-aware reinforcement learning for large reasoning models},
  author={Li, Junyi and Ng, Hwee Tou},
  journal={arXiv preprint arXiv:2505.24630},
  year={2025}
}

@article{kulkarni2025evaluating,
  title={Evaluating evaluation metrics--the mirage of hallucination detection},
  author={Kulkarni, Atharva and Zhang, Yuan and Moniz, Joel Ruben Antony and Ge, Xiou and Tseng, Bo-Hsiang and Piraviperumal, Dhivya and Swayamdipta, Swabha and Yu, Hong},
  journal={arXiv preprint arXiv:2504.18114},
  year={2025}
}

@article{orgad2024llms,
  title={Llms know more than they show: On the intrinsic representation of llm hallucinations},
  author={Orgad, Hadas and Toker, Michael and Gekhman, Zorik and Reichart, Roi and Szpektor, Idan and Kotek, Hadas and Belinkov, Yonatan},
  journal={arXiv preprint arXiv:2410.02707},
  year={2024}
}

@article{bar2025beyond,
  title={Beyond Token Probes: Hallucination Detection via Activation Tensors with ACT-ViT},
  author={Bar-Shalom, Guy and Frasca, Fabrizio and Galron, Yaniv and Ziser, Yftah and Maron, Haggai},
  journal={arXiv preprint arXiv:2510.00296},
  year={2025}
}

@article{zhang2025incentivizing,
  title={Incentivizing LLMs to Self-Verify Their Answers},
  author={Zhang, Fuxiang and Xu, Jiacheng and Wang, Chaojie and Cui, Ce and Liu, Yang and An, Bo},
  journal={arXiv preprint arXiv:2506.01369},
  year={2025}
}

@inproceedings{sawczyn2026factselfcheck,
  title={Factselfcheck: Fact-level black-box hallucination detection for llms},
  author={Sawczyn, Albert and Binkowski, Jakub and Janiak, Denis and Gabrys, Bogdan and Kajdanowicz, Tomasz Jan},
  booktitle={Findings of the Association for Computational Linguistics: EACL 2026},
  pages={5603--5621},
  year={2026}
}

\newpage
\appendix

\section{Implementation Details}
\label{app:implementation}
In our experiments, we implement the MAVEN framework using the following settings to ensure reproducibility. For the main evaluation, we utilize GPT-4o accessed via API. The decoding temperature is strictly set to 0.0 to ensure deterministic outputs for reasoning tasks, and the maximum generation length is set to 4096 tokens.

Regarding the framework hyperparameters, we set the default maximum deliberation loops $\mathcal{K}_{max}$ to $3$. For the Judge Agent, the acceptance threshold for the weighted average score is set to $4.2$, and the strict factual accuracy threshold is set to $4.0$. 

\lstset{
    basicstyle=\ttfamily\footnotesize,
    breaklines=true,
    breakatwhitespace=false,
    columns=flexible,
    keepspaces=true,
    tabsize=4,
    resetmargins=true,
    xleftmargin=0pt,
    xrightmargin=0pt,
    frame=none,
}

\newtcolorbox{promptbox}[2][]{%
  enhanced,
  breakable,
  colback=white,
  colframe=black!95,
  title={#2},
  fonttitle=\bfseries,
  coltitle=white,
  colbacktitle=black!95,
  boxrule=1.0pt,
  arc=2pt,
  left=6pt,right=6pt,top=6pt,bottom=6pt,
  before skip=8pt, after skip=8pt,  
  #1
}

\section{Complete Prompt Templates}
\label{app:prompts}
In this section, we provide the exact system and user prompt templates used by each agent within the MAVEN framework. Variables dynamically injected during runtime are denoted by \texttt{\{variable\_name\}}.

\subsection{Router Agent}
The Router Agent analyzes the complexity of the user query to determine whether to use the FAST\_PATH or FULL\_MAVEN deliberation process.

\begin{promptbox}[colback=blue!5,colframe=blue!35!gray,colbacktitle=blue!55]{System Prompt: Router}
\begin{lstlisting}
You are a highly efficient query analysis agent. Your primary task is to estimate the 'reasoning hops' required to fully answer a user's query and then decide on the appropriate processing path.
**Definition of Reasoning Hops:**
- **1 Hop:** A direct, single-fact lookup. The answer is immediately available from a knowledge base.
- **2 Hops:** Requires a simple inference or the combination of two distinct facts. There is one intermediate logical step.
- **> 2 Hops (Complex):** Requires multi-step analysis, synthesis of different pieces of information, comparison of pros and cons, planning, or addressing subjective/controversial topics.
**Decision Criteria:**
- **FAST_PATH**: Use for queries requiring **2 or fewer** reasoning hops. These are relatively straightforward.
- **FULL_MAVEN**: Use for queries requiring **more than 2** reasoning hops. These are complex and benefit from a full deliberation process.
**Examples:**
- Query: "Who wrote 'Hamlet'?" -> **1 Hop**. Direct fact. -> **FAST_PATH**.
- Query: "What is the capital of the country where the Eiffel Tower is located?" -> **2 Hops**. (1. Eiffel Tower -> France; 2. Capital of France -> Paris). -> **FAST_PATH**.
- Query: "Analyze the pros and cons of nuclear energy." -> **> 2 Hops**. (Requires listing pros, listing cons, weighing them, structuring the analysis). -> **FULL_MAVEN**.
- Query: "A winter storm is ravaging a city... While driving, he discovers..." -> **> 2 Hops**. (Requires analyzing the context, inferring conditions, evaluating options, and eliminating distractors). -> **FULL_MAVEN**.
Your output MUST be a single, strict JSON object with three keys: "decision", "estimated_hops", and "reason".
\end{lstlisting}
\end{promptbox}

\begin{promptbox}[colback=blue!5,colframe=blue!35!gray,colbacktitle=blue!55]{User Prompt: Router}
\begin{lstlisting}
Please analyze the following user query, estimate the reasoning hops required, and classify it.
User Query: "{query}"
Strictly output your decision in the following JSON format:
{
    "decision": "...",
    "estimated_hops": "...",
    "reason": "..."
}
\end{lstlisting}
\end{promptbox}

\subsection{Fast Proposer Agent}
\begin{promptbox}[colback=NavyBlue!5,colframe=NavyBlue!35!gray,colbacktitle=NavyBlue!55]{System Prompt: Fast Proposer}
\begin{lstlisting}
You are a decisive expert assistant. Your task is to provide a direct, intuitive answer to the question. Focus on being factually correct and provide a confidence score.
\end{lstlisting}
\end{promptbox}

\begin{promptbox}[colback=NavyBlue!5,colframe=NavyBlue!35!gray,colbacktitle=NavyBlue!55]{User Prompt: Fast Proposer}
\begin{lstlisting}
Question: "{query}"
Please provide:
1. Your direct answer (if multiple choice, state the option and text).
2. A brief, 1-sentence reason for your choice.
3. Your confidence score between 0.0 and 1.0.

Format your response as a JSON object:
{
    "answer": "...",
    "reason": "...",
    "confidence": 0.95
}
\end{lstlisting}
\end{promptbox}

\subsection{Planner Agent}
\begin{promptbox}[colback=violet!5,colframe=violet!35!gray,colbacktitle=violet!55]{System Prompt: Planner}
\begin{lstlisting}
You are a strategic designer. Your task is to create a high-level, logically clear execution plan to answer the user's question. The plan should be broken down into several logical steps.
\end{lstlisting}
\end{promptbox}

\begin{promptbox}[colback=violet!5,colframe=violet!35!gray,colbacktitle=violet!55]{User Prompt: Planner (Initial Planning)}
\begin{lstlisting}
User Question: "{query}"
Initial Intuitive Answers from experts:
{fast_refs}
Please create a high-level plan for this question. The plan should include a goal and a series of clear steps to guide the subsequent information gathering and drafting process. Use the initial answers as a reference to identify which areas need the most rigorous verification.
The plan should be output in Markdown format.
\end{lstlisting}
\end{promptbox}

\begin{promptbox}[colback=violet!5,colframe=violet!35!gray,colbacktitle=violet!55]{User Prompt: Planner (Replanning)}
\begin{lstlisting}
Original User Question: "{query}"
The previous plan has fundamental flaws and needs to be replanned.
Feedback from the "Judge": "{feedback}"
Based on the feedback above, please generate a new, revised high_level_plan. Ensure the new plan addresses all the issues pointed out in the feedback.
The plan should be output in Markdown format.
\end{lstlisting}
\end{promptbox}

\subsection{Multi-Proposer Agent}
The Proposer Agent generates drafts based on the plan. We introduce diversity by appending specific instructions to the base prompt.

\begin{promptbox}[colback=orange!5,colframe=orange!35!gray,colbacktitle=orange!55]{System Prompt: Proposer}
\begin{lstlisting}
You are an informative drafter. Your task is to follow the provided plan to quickly and broadly generate an information-rich preliminary draft answer.
\end{lstlisting}
\end{promptbox}

\begin{promptbox}[colback=orange!5,colframe=orange!35!gray,colbacktitle=orange!55]{User Prompt: Base Proposer}
\begin{lstlisting}
Original User Question:
---
{query}
---
Initial Intuitive Answers for reference:
{fast_refs}
Execution Plan:
---
{plan}
---
Please generate a detailed preliminary draft based on the original user question, the execution plan, and the initial expert intuition provided above. The draft should be well-structured, comprehensive, and directly address the user's question.
\end{lstlisting}
\end{promptbox}

To enforce draft diversity, we append one of the following instructions to the base prompt for parallel drafting:
\begin{itemize}[nosep, leftmargin=1.5em]
    \item \textbf{Skeptical Perspective:} \texttt{Please pay special attention that your draft needs to take a critical and skeptical perspective, proactively exploring potential risks and negative factors.}
    \item \textbf{Quantitative Perspective:} \texttt{Please pay special attention that your draft needs to be centered on data and quantitative analysis, supporting its points with numbers and statistical results as much as possible.}
\end{itemize}

\subsection{Synthesizer Agent}
\begin{promptbox}[colback=cyan!5,colframe=cyan!35!gray,colbacktitle=cyan!55]{System Prompt: Synthesizer}
\begin{lstlisting}
You are a senior information integration specialist. Your task is to merge drafts from multiple sources into a single, comprehensive, balanced, and well-structured consensus draft.
\end{lstlisting}
\end{promptbox}

\begin{promptbox}[colback=cyan!5,colframe=cyan!35!gray,colbacktitle=cyan!55]{User Prompt: Synthesizer}
\begin{lstlisting}
Original Execution Plan:
---
{plan}
---
Below are different versions of drafts written by multiple drafters based on the same plan.
{drafts_string}
Your task is to:
1.  **Identify Consensus**: Find the core information that is mentioned in all drafts and on which they agree.
2.  **Identify Differences and Contradictions**: Find the contradictions or differences in viewpoints that exist between the different drafts.
3.  **Identify Unique Information**: Find information that is unique to one draft but valuable for answering the question.
4.  **Synthesize**: Based on the analysis above, generate a new, single "consensus draft". This draft should:
    *   Contain the consensus and valuable unique information from all sources.
    *   For contradictions, state them objectively, or choose the side with stronger evidence.
    *   Be well-structured, logically coherent, and fully respond to the original plan.
Please output this final consensus draft.
\end{lstlisting}
\end{promptbox}

\subsection{Skeptic Agent}
\begin{promptbox}[colback=red!5,colframe=red!35!gray,colbacktitle=red!55]{System Prompt: Skeptic}
\begin{lstlisting}
You are an extremely rigorous critical reviewer. Your task is to systematically challenge a draft. Your questions will be answered by another powerful AI model, so please ask questions that test its depth of knowledge and reasoning abilities.
Your output must be in strict JSON format.
\end{lstlisting}
\end{promptbox}

\begin{promptbox}[colback=red!5,colframe=red!35!gray,colbacktitle=red!55]{User Prompt: Skeptic}
\begin{lstlisting}
Please conduct a rigorous internal review of the following draft. Your questions will be answered by another powerful AI, so formulate them to probe for depth, context, and logical consistency.
Draft:
---
{draft}
---
Please follow these review instructions:
1.  **Fact Verification**: For key claims (like numbers, stats, events), ask for verification and context.
    -   New example: "Please verify and elaborate on the statement 'unemployment fell by 5%', including the time period and methodology."
2.  **Logical Coherence**: Identify unclear or contradictory points in the argument and ask for clarification.
    -   Example: "The draft first states the market is growing, but later says profits are declining. Please explain the potential relationship between these two seemingly contradictory trends."
3.  **Causal Scrutiny**: Question if a stated cause-and-effect relationship is oversimplified. Ask for a deeper explanation.
    -   Example: "The draft claims A led to B. What other significant factors, such as C and D, might have also contributed to B? Please elaborate."
4.  **Adversarial Questioning**: Pose questions that challenge the draft's core assumptions or seek counterexamples.
    -   Example: "The draft's central argument is X. Under what conditions would this argument not hold true? Please provide a specific counterexample or edge case."
Your output must be a JSON array, where each object contains four keys: "claim" (the original statement being questioned), "question" (your verifiable question), "type" (one of: "Factual", "Logical", "Causal", or "Adversarial"), and "priority" (one of: "High", "Medium", "Low").
Ensure your questions are self-contained and clear, so another AI can understand and answer them directly.
\end{lstlisting}
\end{promptbox}

\subsection{Responder Agent}
\begin{promptbox}[colback=brown!5,colframe=brown!35!gray,colbacktitle=brown!55]{System Prompt: Responder}
\begin{lstlisting}
You are a knowledgeable author. You have just completed the draft below. Now, a reviewer has raised some questions about your draft. Please defend and explain these points based solely on the knowledge you had when writing the draft.
\end{lstlisting}
\end{promptbox}

\begin{promptbox}[colback=brown!5,colframe=brown!35!gray,colbacktitle=brown!55]{User Prompt: Responder}
\begin{lstlisting}
Your Draft:
---
{draft}
---
Reviewer's Questions:
---
{questions_string}
---
Please answer the above questions one by one **without conducting any new external information retrieval**. Your goal is to explain the rationale behind your views and statements.
Please output your answers in a clear list format.
\end{lstlisting}
\end{promptbox}

\subsection{Judge Agent}
\begin{promptbox}[colback=magenta!5,colframe=magenta!35!gray,colbacktitle=magenta!55]{System Prompt: Judge}
\begin{lstlisting}
You are a deliberate decision-making center. Your core task is to synthesize all information to make a fair, explainable, and quantifiable judgment, and to decide the direction of the process.
Your thought process should follow these steps, but your final output must be a single, strict JSON object:
1.  **Conflict Analysis**: Compare the evidence report with the current draft, identifying inconsistencies.
2.  **Logical Evaluation**: Check if the draft's internal chain of logic is sound and free of contradictions.
3.  **Counterfactual Evaluation**: Negate the core conclusion to see if it sharply contradicts the evidence, thereby testing causal robustness.
4.  **Formulate Revision Plan**: If the draft has issues, mentally construct a plan for revision.
5.  **Generate Revised Draft**: Based on your plan, generate an improved version of the draft. If no changes are needed, use the original.
6.  **Quantitative Assessment & Decision**: Score the draft on dimensions like factual accuracy and logical validity. Based on the scores and thresholds (Accept Threshold: {accept_threshold}, Fact Threshold: {fact_threshold}), make a final decision (ACCEPT, REJECT, REPLAN).
\end{lstlisting}
\end{promptbox}

\begin{promptbox}[colback=magenta!5,colframe=magenta!35!gray,colbacktitle=magenta!55]{User Prompt: Judge}
\begin{lstlisting}
Based on your system instructions, conduct a comprehensive evaluation of the following information.
**Original Plan:**
---
{plan}
---
**Current Draft:**
---
{draft}
---
**Evidence Report (from Researcher):**
---
{evidence_report}
---
Strictly output your final verdict in the following JSON format. Do not include any extra text, explanations, or Markdown formatting outside the JSON structure.
{
    "reasoning": "Provide your complete, step-by-step thought process here. Clearly explain how you evaluated the facts, logic, and causal relationships to arrive at your final decision.",
    "revised_draft": "Place the final, revised version of the draft here based on your evaluation. If the original draft is accepted, simply copy the original draft here.",
    "final_decision": "Write only one of three decisions here: 'ACCEPT', 'REJECT', or 'REPLAN'"
}
\end{lstlisting}
\end{promptbox}

\subsection{Stylist Agent}
\begin{promptbox}[colback=teal!5,colframe=teal!35!gray,colbacktitle=teal!55]{System Prompt: Stylist}
\begin{lstlisting}
You are a final copy editor. Your core objective is to improve the fluency and readability of the final manuscript without altering any facts or logic.
\end{lstlisting}
\end{promptbox}

\begin{promptbox}[colback=teal!5,colframe=teal!35!gray,colbacktitle=teal!55]{User Prompt: Stylist}
\begin{lstlisting}
Please polish and format the final draft below.
Ensure that no facts, data, or core logic are changed. Only improve the quality of the language and expression.
Draft:
---
{draft}
---
Please output the polished final version.
\end{lstlisting}
\end{promptbox}

\section{Detailed Experimental Setup and Reproducibility}
\label{sec:appendix_exp_details}

In this section, we provide comprehensive implementation details, hyperparameter configurations, and the specific rubrics used for our LLM-as-a-Judge evaluation metrics to ensure the full reproducibility of our experiments.

\subsection{Implementation Details and Hyperparameters}
MAVEN is implemented via a modular blackboard architecture, enabling structured, coordinated interaction between specialized reasoning agents. Unless otherwise specified, DeepSeek-V3.2 is utilized as the default backbone across all agentic roles. To balance reasoning depth against inference latency and API costs, the maximum number of deliberation iterations is strictly capped at $T_{max}=3$. Furthermore, the generation temperature is fixed at $0.0$ for all reasoning, proposing, and evaluation agents to ensure deterministic behavior and minimize stochastic variance during benchmarking. The complete set of agentic network configurations, capacity constraints, and judicial decision thresholds are summarized in Table~\ref{tab:hyperparams_appendix}.

\begin{table}[h]
  \caption{Global hyperparameters and decision thresholds used in the MAVEN pipeline.}
  \label{tab:hyperparams_appendix}
  \centering 
  \begin{tabular}{llc} 
    \toprule
    \textbf{Category} & \textbf{Parameter} & \textbf{Value} \\
    \midrule
    \multirow{2}{*}{Deliberation} & Max Iterations ($T_{max}$) & 3 \\
                                  & Num. FastProposers ($N$) & 5 \\
    \midrule
    \multirow{2}{*}{Judge Thresholds} & Accept Score Threshold ($\alpha$) & 4.2 \\
                                      & Fact Score Threshold ($\beta$) & 4.0 \\
    \midrule
    \multirow{3}{*}{Model Params} & Primary LLM Backbone & DeepSeek-V3.2 \\
                                  & Max Generation Tokens & 4096 \\
                                  & Logic/Evaluation Temperature & 0.0 \\
    \bottomrule
  \end{tabular}
\end{table}

\subsection{Fine-grained Evaluation Metrics}
In complex multi-hop reasoning and verification tasks, binary accuracy alone fails to capture the nuances of argument construction, hallucination mitigation, and causal validity. Consequently, we employ a GPT-4o-based expert judge to evaluate the intermediate and final justifications on a continuous $1-100$ scale. To maintain strict evaluation consistency and prevent the "LLM-as-a-Judge" from exhibiting random biases, all metrics are formally defined via structured rubrics as follows:

\begin{promptbox}[colback=ForestGreen!5,colframe=ForestGreen!35!gray,colbacktitle=ForestGreen!55]{Justification Plausibility \& Causal Depth Score}
\begin{lstlisting}
*   **Objective:** Assess the logical flow that bridges the technical principles to the final answer.
*   **Rubric:**
    *   **1-49 (Superficial/Flawed):** The technical reasoning is broken, contradictory, or jumps to conclusions without proper engineering justification.
    *   **50-69 (Plausible):** Provides a basic technical connection that supports the conclusion.
    *   **70-89 (Well-Reasoned):** Presents a **clear, step-by-step technical reasoning chain**. It explicitly connects the relevant equipment parameters, operational principles, or maintenance procedures required to answer the question.
    *   **90-100 (Deeply Principled):** Flawless technical execution that breaks down the question into its core engineering principles before synthesizing the final answer with proper causal relationships.
\end{lstlisting}
\end{promptbox}

\begin{promptbox}[colback=RoyalBlue!5,colframe=RoyalBlue!35!gray,colbacktitle=RoyalBlue!55]{Factual Accuracy \& Contextualization Score}
\begin{lstlisting}
*   **Objective:** Assess if the technical facts, parameters, and domain knowledge the AI retrieves are accurate.
*   **Rubric:**
    *   **1-49 (Major Errors):** Contains significant technical errors regarding equipment specifications, operational parameters, or maintenance procedures.
    *   **50-69 (Accurate but Weak):** Technical facts are correct but poorly connected to the specific equipment or scenario in the question.
    *   **70-89 (Accurate & Relevant):** Retrieves the correct technical knowledge required to answer the specific question about UHV equipment.
    *   **90-100 (Flawless & Contextualized):** Not only retrieves the correct technical facts but contextualizes them perfectly within the specific station, equipment type, and operational scenario mentioned in the question.
\end{lstlisting}
\end{promptbox}

\begin{promptbox}[colback=Plum!5,colframe=Plum!35!gray,colbacktitle=Plum!55]{Completeness \& Alternative Analysis Score}
\begin{lstlisting}
*   **Objective:** Assess if the answer addresses all aspects of the question and considers relevant operational considerations.
*   **Rubric:**
    *   **1-49 (Incomplete):** Misses major technical points or key aspects of the maintenance/operation question.
    *   **50-69 (Sufficient):** Covers the main technical points but ignores important operational details or safety considerations.
    *   **70-89 (Comprehensive):** Thoroughly addresses all aspects of the prompt including equipment specifics, monitoring parameters, and operational procedures.
    *   **90-100 (Expert Analysis):** Meets the "Comprehensive" standard AND provides additional insights on related considerations, potential issues, or cross-references to relevant standards/regulations.
\end{lstlisting}
\end{promptbox}

\begin{promptbox}[colback=Aquamarine!5,colframe=Aquamarine!35!gray,colbacktitle=Aquamarine!55]{Argument Richness \& Structure Score}
\begin{lstlisting}
*   **Objective:** Quantify the structural quality and technical depth.
*   **Additive Scoring Rubric:**
    *   **Start with a baseline score of 50** for any coherent, paragraph-level technical explanation.
    *   **+15 Points** for a **clear, organized structure** (e.g., categorized by inspection items, monitoring parameters, or procedural steps).
    *   **+15 Points** for explicitly **identifying the key technical parameters or thresholds** relevant to the question.
    *   **+10 Points** for **technical nuance or consideration of edge cases/abnormal conditions**.
    *   **+10 Points** for **specific equipment references, numerical specifications, or industry standard citations**.
    *   **Final Score is the sum, capped at 100.**
\end{lstlisting}
\end{promptbox}

\section{Iteration-Budget Sensitivity Analysis}
\label{sec:appendix_full_results}

Table~\ref{tab:iteration_sensitivity_full_data} reports the raw scores underlying the iteration-budget sensitivity curves in Section~\ref{subsec:sensitivity}. We vary the maximum number of deliberation rounds $T_{\max}$, where $T_{\max}=1$ corresponds to a single-pass generation without iterative refinement. Overall, the improvements concentrate in early iterations (typically up to $T_{\max}\le 3$), after which performance saturates and remains stable. This behavior indicates that MAVEN’s stopping mechanism avoids unnecessary late-stage revisions and mitigates drift at larger iteration budgets.

\begin{table}[]
\centering
\caption{Iteration-budget sensitivity (raw numbers). MAVEN performance under different maximum iteration budgets $T_{\max}$ on OBQA, TQA, HaluEval, and SQA. $T_{\max}=1$ denotes single-pass generation without deliberation. Bold marks the best value per metric within each dataset. All values are in \%.}
\label{tab:iteration_sensitivity_full_data}
\resizebox{\textwidth}{!}{
\begin{tabular}{c | ccccc | ccccc}
\toprule
\multirow{2}{*}{\textbf{$T_{max}$}} & \multicolumn{5}{c|}{\textbf{OpenBookQA (OBQA)}} & \multicolumn{5}{c}{\textbf{TruthfulQA (TQA)}} \\
\cmidrule(lr){2-6} \cmidrule(lr){7-11}
& \textbf{Acc} & \textbf{JCD} & \textbf{F\&C} & \textbf{C\&A} & \textbf{ARS} 
& \textbf{Acc} & \textbf{JCD} & \textbf{F\&C} & \textbf{C\&A} & \textbf{ARS} \\
\midrule
1 & 93.33 & 90.27 & 91.05 & 88.21 & 93.85 & 87.00 & 88.95 & 88.18 & 88.55 & 92.20 \\
2 & 95.67 & 92.44 & 93.10 & 89.80 & 95.20 & 88.33 & 90.35 & 89.70 & 89.20 & 95.20 \\
3 & 96.33 & 92.92 & \textbf{93.77} & 89.92 & 96.10 & 90.00 & 90.77 & 90.90 & 89.37 & 96.38 \\
4 & 96.00 & 92.97 & 93.20 & \textbf{90.71} & 96.90 & 91.00 & 90.95 & \textbf{92.75} & \textbf{91.26} & 96.90 \\
5 & 96.67 & 93.19 & 92.75 & 90.28 & 96.70 & 91.67 & 91.05 & 92.65 & 91.16 & 96.94 \\
6 & \textbf{97.00} & \textbf{93.28} & 93.33 & 90.43 & \textbf{97.60} & \textbf{92.00} & \textbf{91.06} & 92.71 & 91.72 & \textbf{97.25} \\
\midrule
\midrule
\multirow{2}{*}{\textbf{$T_{max}$}} & \multicolumn{5}{c|}{\textbf{HaluEval}} & \multicolumn{5}{c}{\textbf{StrategyQA (SQA)}} \\
\cmidrule(lr){2-6} \cmidrule(lr){7-11}
& \textbf{Acc} & \textbf{JCD} & \textbf{F\&C} & \textbf{C\&A} & \textbf{ARS} 
& \textbf{Acc} & \textbf{JCD} & \textbf{F\&C} & \textbf{C\&A} & \textbf{ARS} \\
\midrule
1 & 97.33 & 89.99 & 97.43 & 90.49 & 90.06 & 76.67 & 89.45 & 90.75 & 86.03 & 90.10 \\
2 & 98.00 & 90.50 & 97.86 & 90.90 & 90.65 & 80.00 & 91.05 & 91.35 & 87.73 & 93.05 \\
3 & 98.33 & 90.70 & 98.08 & 91.27 & 90.85 & 82.00 & 91.46 & 92.72 & 87.84 & 93.11 \\
4 & 98.67 & 90.55 & 98.25 & 91.45 & 90.70 & 82.33 & 91.50 & 92.50 & 87.88 & 92.89 \\
5 & \textbf{99.00} & \textbf{90.91} & 98.35 & 91.55 & 90.95 & \textbf{82.67} & 91.75 & 92.85 & 87.99 & 93.05 \\
6 & 98.33 & 90.85 & \textbf{98.50} & \textbf{91.90} & \textbf{91.32} & \textbf{82.67} & \textbf{91.83} & \textbf{93.43} & \textbf{88.22} & \textbf{93.65} \\
\bottomrule
\end{tabular}
}
\end{table}

\section{Case Studies}
\label{app:case_study}
Table \ref{tab:case_study_full} demonstrates a complete workflow of the MAVEN framework on a complex reasoning question from StrategyQA.

\begin{longtable}{p{0.15\textwidth} p{0.82\textwidth}}
\caption{MAVEN Case Study (Full Deliberation Trace). Complete operational trace demonstrating the rigorous fact-checking and causal refinement on a StrategyQA query. The system autonomously escalates the reasoning from a basic hobbyist level to an advanced biological and ecological comparison.}
\label{tab:case_study_full} \\
\toprule
\multicolumn{2}{l}{\textbf{User Query:} \textit{Are goldfish more difficult to care for than isopods?}} \\
\multicolumn{2}{l}{\textbf{Ground Truth:} \textit{True (Yes)}} \\
\midrule
\endfirsthead

\multicolumn{2}{c}
{{\bfseries Table \thetable\ continued from previous page}} \\
\toprule
\endhead

\midrule \multicolumn{2}{r}{{Continued on next page}} \\ \midrule
\endfoot

\bottomrule
\endlastfoot
\multicolumn{2}{l}{\textbf{ITERATION 1: Initial Synthesis Draft (Accepted for Review)}} \\
\textbf{Initial Draft} & 
\begin{itemize}[nosep, leftmargin=1.5em]
    \item \textbf{Environmental Control:} Goldfish require large aquaria (20--55+ gal) to stabilize the mandatory nitrogen cycle and maintain 0 ppm Ammonia.
    \item \textbf{Maintenance:} Goldfish are labor-intensive (weekly water changes, 1-2 hours). Isopods are minimal and observational (<20 mins).
    \item \textbf{Conclusion:} Yes, goldfish are unequivocally more difficult due to required precision and small margin for error.
\end{itemize} \\
\midrule

\multicolumn{2}{l}{\textbf{FORENSIC SCRUTINY (\textsc{Skeptic} Probes)}} \\
\textbf{Q1 (Factual)} & "Verify the basis for the cited 20+ gal minimum. What physiological data supports these volume requirements for cold-water species?" \\
\textbf{Q2 (Causal)} & "Explain the precise biochemical mechanisms of the nitrogen cycle ($NH_3 \rightarrow NO_2^- \rightarrow NO_3^-$). What microbial genera catalyze each step? Quantify the difference in microbial complexity compared to isopod decomposition." \\
\textbf{Q3 (Logical)} & "Compare the technical demands of maintaining a humidity gradient (VPD) versus water testing. What are the specific physiological consequences of deviations?" \\
\textbf{Q4 (Adversarial)} & "Is there a coherent framework where isopods are 'more difficult'? Consider the specialized material sourcing (pesticide-free leaf litter) and deep ecological knowledge required." \\
\midrule

\multicolumn{2}{l}{\textbf{PARAMETRIC PROBING \& GROUNDING (\textsc{Researcher} Evidence Report)}} \\
\textbf{On Biology} & "The 20-55+ gallon rule reflects the biology of 'Tank Busters': indeterminate growth, high metabolic waste, and high oxygen demand for cold-water cyprinids." \\
\textbf{On Ecology} & "The aquarium nitrogen cycle is a **linear, specialized chemoautotrophy** dominated by a few genera (*Nitrosomonas*, *Nitrospira*). It is functionally **fragile**. The isopod enclosure is a **web-like, generalized heterotrophy** involving diverse kingdoms (Fungi, Bacteria) and is highly **resilient** due to functional redundancy." \\
\textbf{On Difficulty} & "A valid alternative definition of difficulty exists: 'Logistical/Knowledge-Based Difficulty'. Sourcing specific, uncontaminated decaying wood and establishing bioactive microfauna is a non-trivial hurdle underweighted in the initial draft." \\
\midrule

\multicolumn{2}{l}{\textbf{ADJUDICATION (\textsc{Judge} Verdict)}} \\
\textbf{Verdict} & \textbf{\textsc{REJECT}} \\
\textbf{Reasoning} & "The evidence report validates the technical claims but highlights substantive gaps: (1) It provides a biochemical depth (fragile linear autotrophy vs. resilient heterotrophy) missing from the draft. (2) It identifies that logistical/sourcing challenges for isopods represent a valid dimension of difficulty that is underweighted. The draft must be revised to integrate these scientific specifics and acknowledge the logistical difficulty dimension." \\
\midrule

\multicolumn{2}{l}{\textbf{ITERATION 2: Revised Draft \& Final Verdict (\textsc{Accept})}} \\
\textbf{Revised Text} & 
\begin{itemize}[nosep, leftmargin=1.5em]
    \item \textit{(Integration of Biology)} "Must establish a biological nitrogen cycle—a **simplified, vulnerable metabolic pathway** where specialized chemoautotrophic bacteria sequentially oxidize toxic ammonia. In contrast, isopods rely on a **complex, resilient heterotrophic decomposition network**..."
    \item \textit{(Integration of Logistics)} "Isopod care introduces distinct, non-trivial challenges in material sourcing (pesticide-free leaf litter) and ecological balancing, but its ongoing parameter management is fundamentally simpler."
    \item \textit{(Final Conclusion)} "\textbf{Yes, goldfish are significantly more difficult...} when evaluated on procedural criteria and acute error margin. It is noted that alternative definitions focusing on logistical sourcing could lead to different judgments."
\end{itemize} \\
\end{longtable}

\section{Licenses for Existing Assets}

We properly credit all datasets and models used in this paper, and we respect all applicable licenses and terms of use.

We use four publicly available datasets in our experiments. OpenBookQA is released under the Apache License 2.0. TruthfulQA is released under the Apache-2.0 license. HaluEval is released under the MIT License. StrategyQA is released under the MIT License and permitted for research use. All datasets were accessed through their official repositories or Hugging Face.

For large language models, we accessed GPT series via the official OpenAI API, Gemini via the Google Gemini API, and DeepSeek via the official DeepSeek Open Platform API. All API usages strictly comply with the respective providers' Terms of Use and research publication policies. These models were used exclusively for inference and evaluation purposes. We did not perform any fine-tuning, distillation, or weight redistribution, nor did we create or re-package any datasets.

No terms of service were violated in this work.

\end{document}